\documentclass[twoside]{article}

\usepackage[accepted]{aistats2025}

\usepackage{hyperref}
\usepackage{url}
\usepackage{mathtools}
\DeclareMathAlphabet{\mymathbb}{U}{BOONDOX-ds}{m}{n}

\usepackage{graphicx}
\usepackage{subfig}
\usepackage{multirow}

\usepackage{float}
\floatstyle{plaintop}
\restylefloat{table}

\usepackage[table]{xcolor}
\usepackage{pgfplots}
\usepackage{booktabs}
\usepackage{diagbox}

\usepackage{amsmath,amssymb,amsthm}
\usepackage{bm}

\newtheorem{thm}{Theorem}
\newtheorem{assumption}{Assumption}

\begin{document}
\abovedisplayskip=4pt
\abovedisplayshortskip=0pt
\belowdisplayskip=3pt
\belowdisplayshortskip=0pt

\twocolumn[
\aistatstitle{Global Ground Metric Learning with Applications to scRNA data}
\aistatsauthor{Damin Kühn \And Michael T. Schaub }
\aistatsaddress{Department of Computer Science \\ RWTH Aachen University, Germany}]

\begin{abstract}
Optimal transport provides a robust framework for comparing probability distributions. Its effectiveness is significantly influenced by the choice of the underlying ground metric. Traditionally, the ground metric has either been (i) predefined, e.g., as the Euclidean distance, or (ii) learned in a supervised way, by utilizing labeled data to learn a suitable ground metric for enhanced task-specific performance.
Yet, predefined metrics typically cannot account for the inherent structure and varying importance of different features in the data, and existing supervised approaches to ground metric learning often do not generalize across multiple classes or are restricted to distributions with shared supports.   
To address these limitations, we propose a novel approach for learning metrics for arbitrary distributions over a shared metric space. 
Our method provides a distance between individual points like a global metric, but requires only class labels on a distribution-level for training. The learned global ground metric enables more accurate optimal transport distances, leading to improved performance in embedding, clustering and classification tasks. We demonstrate the effectiveness and interpretability of our approach using patient-level scRNA-seq data spanning multiple diseases. 
\end{abstract}

\section{INTRODUCTION}
Optimal transport (OT) has reemerged as a powerful mathematical framework for comparing probability distributions. Recently, it has found applications in analyzing data across various domains like computer vision ~\cite{bonneel2023survey}, natural language processing ~\cite{sato2022re} and computational genomics \cite{schiebinger2019optimal,joodaki2024detection}.

Our work is motivated by patient-level single-cell RNA (scRNA) analysis, where the available data consists of the expression patterns of individual cells from different patients.
Specifically, we consider a setting in which we have multiple patients, each represented by a set of cells, where each cell is characterized by a high-dimensional gene expression vector.
Hence, these gene expression vectors of cells from the same patient form a distribution over the gene space, i.e., each patient is represented by a distribution over the gene space.
Our objective is to compare these distributions to identify similarities and differences between patients.
Intuitively, OT can be used to probabilistically map two such empirical distributions (i.e., cells from two patients) onto each other while minimizing the overall cost of this transport.
The associated cost of this optimal transport is then a Wasserstein distance $W$, which quantifies the distance (dissimilarity) between the distributions of the two patients.
However, to compute this cost, we need to choose an underlying Ground Metric $d$, which defines the distance between points (gene expression vectors) sampled from the two distributions.
Clearly, the chosen ground metric critically influences the resulting Wasserstein distance $W$.

Yet, most current applications of OT rely on a predefined and fixed ground metric, such as the Euclidean distance. 
Such simple ground metrics are however rarely tailored to the task at hand and may lead to suboptimal results, as they do not account for the inherent structure and varying importance of different features in the data.
Ground Metric Learning (Ground ML) aims at directly learning the distance between sampled points based on prior information about the Wasserstein distance between their distributions. 
A promising application of Ground ML in scRNA data is to leverage different disease states of patients as prior information for their Wasserstein distances. 
Under this setup, Ground ML learns relevant distances between the expression vectors of cells that optimally separate disease states on a patient-level.  
However, early work in Ground ML is constrained to pairs of distributions or requires the distributions to have shared support, which limits its applicability ~\cite{wang2012supervised,cuturi2014ground,huang2016supervised,huizing2022unsupervised}. 

In this paper, we introduce a general framework to ground metric learning for classes of arbitrary distributions over a shared space, called Global Ground Metric Learning (GGML), that circumvents such problems of prior works. Theoretically, GGML can learn arbitrary differentiable metrics $d_\theta$ between sampled points based solely on the class labels of their distributions. 
We demonstrate this with a low rank approximation of a popular learnable global metric, the Mahalanobis distance. 

Notably, our framework enables the joint data analysis on two scales: the level of distributions (patients) and the level of individual points (single cells). Used as a ground metric in OT, the learned metric can improve performance and interpretability of the resulting OT distances in down-stream applications. Used as a global metric, the learned metric offers similar benefits for the analysis of sampled points. This includes applications such as embedding, clustering, classification and feature importance. We validate and benchmark our approach using synthetic and real-world single-cell genomics datasets spanning various diseases. 

\section{BACKGROUND}
\label{sec:background}
\newcommand{\norm}[1]{\left\lVert #1 \right\rVert}
\paragraph{Notation.}
Vectors $\bm{x}$ and matrices $\bm{M}$ are denoted in bold, sets $\mathcal{T}$ calligraphically.  
We use $d(\bm{x}_i,\bm{x}_j)$ to denote a metric $d: \Omega^2 \mapsto \mathbb{R}_{\geq0}$ between two sampled points from the space $\Omega$ (i.e., gene space). 

\paragraph{Mahalanobis distance.}
The Mahalanobis distance is a parameterized, linear metric \cite{davis2007information}. It is defined as: 
\begin{align}
d_M(\bm{x}_i,\bm{x}_j) &= \sqrt{(\bm{x}_i-\bm{x}_j)^T \bm{M} (\bm{x}_i - \bm{x}_j)}  \\
&= \lVert \bm{W}\bm{x}_i-\bm{W}\bm{x}_j \rVert \label{eq:mahala}\\  
s.t. \;\bm{W}^T\bm{W} &= (\bm{Q} \Lambda^{\frac{1}{2}}) (\bm{Q} \Lambda^{\frac{1}{2}})^T =  \bm{Q} \Lambda \bm{Q}^T = \bm{M}  \nonumber
\end{align}

where $\bm{M}$ is a symmetric, positive semi-definite (PSD) matrix that can be learned. 
As every real symmetric PSD matrix $\bm{M}$ can be represented as a product $\bm M = \bm{W^T}\bm{W}$, e.g., using the spectral decomposition, the Mahalanobis distance can be equivalently expressed as performing a linear transformation ($\bm W$) into a subspace in which the Euclidean distance better approximates the desired distances. 

We remark that the Mahalanobis distance is convex and Lipschitz continuous \cite{zantedeschi2016lipschitz}. 
Further, the gradient of $d_M$ with respect to row $\bm{W}_r$ of $\bm{W}$ is given as:
\begin{align}
\frac{\partial d_M}{\partial \bm{W}_r} = \bm{W}_r^T (\bm{x}_i-\bm{x}_j)(\bm{x}_i-\bm{x}_j) \label{eq:mahala_grad}
\end{align}
which allows to efficiently learn an optimal $\bm{W}$ (resp. $\bm{M}$) using gradient descent over some differentiable cost function $L(\bm{W})$
\cite{hocke2014global}. 
\begin{figure}[b!]
    \centering
    \setlength{\tabcolsep}{2pt}
    \begin{tabular}[t]{cc}
        (a) \textbf{Global ML} & (b) \textbf{Global Ground ML}\\ 
        \includegraphics[trim=0.0cm 0cm 0.cm 10.7cm, clip, width=0.49\linewidth]{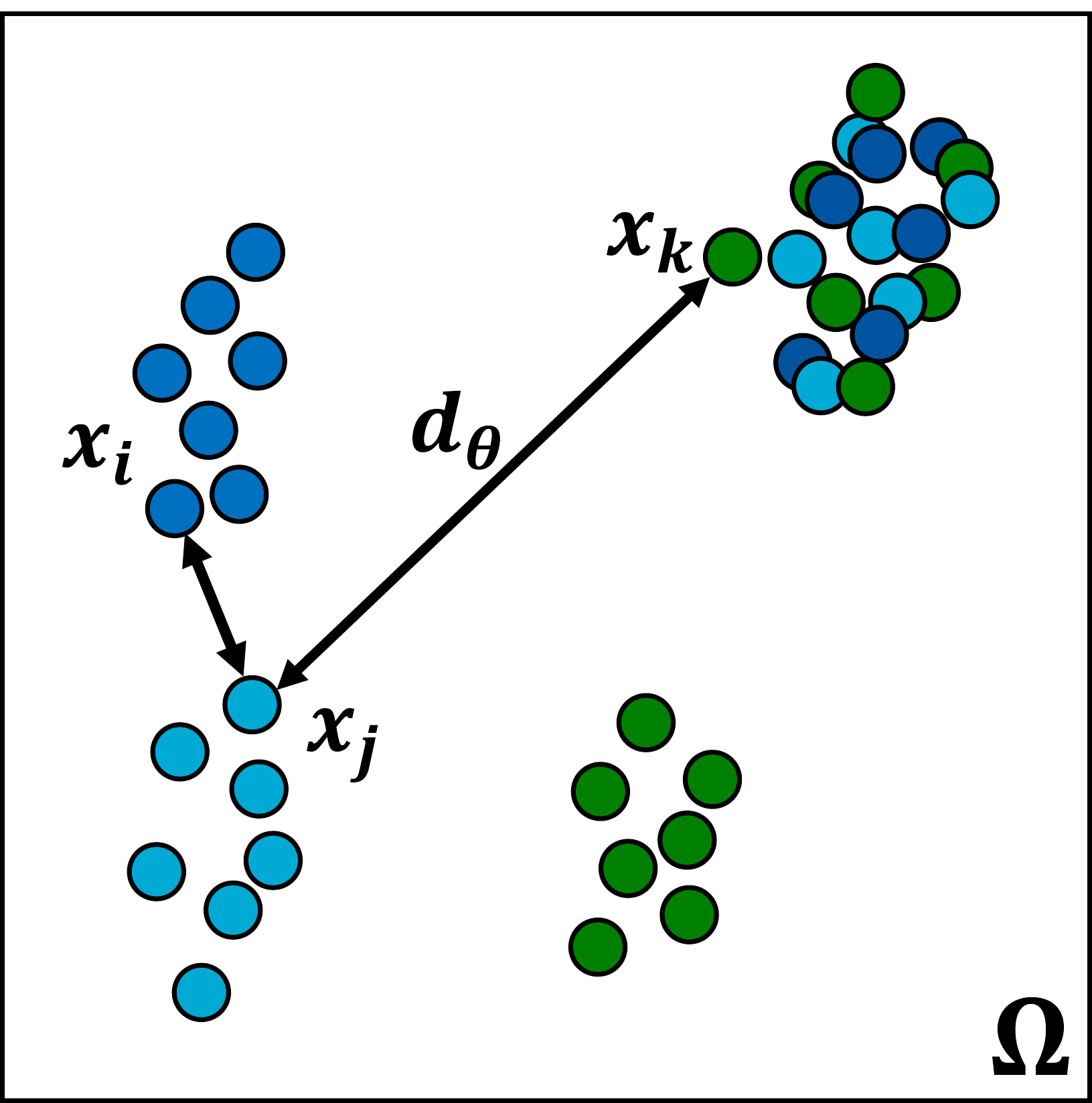} & \includegraphics[trim=0.0cm 0cm 0.cm 10.7cm, clip, width=0.49\linewidth]{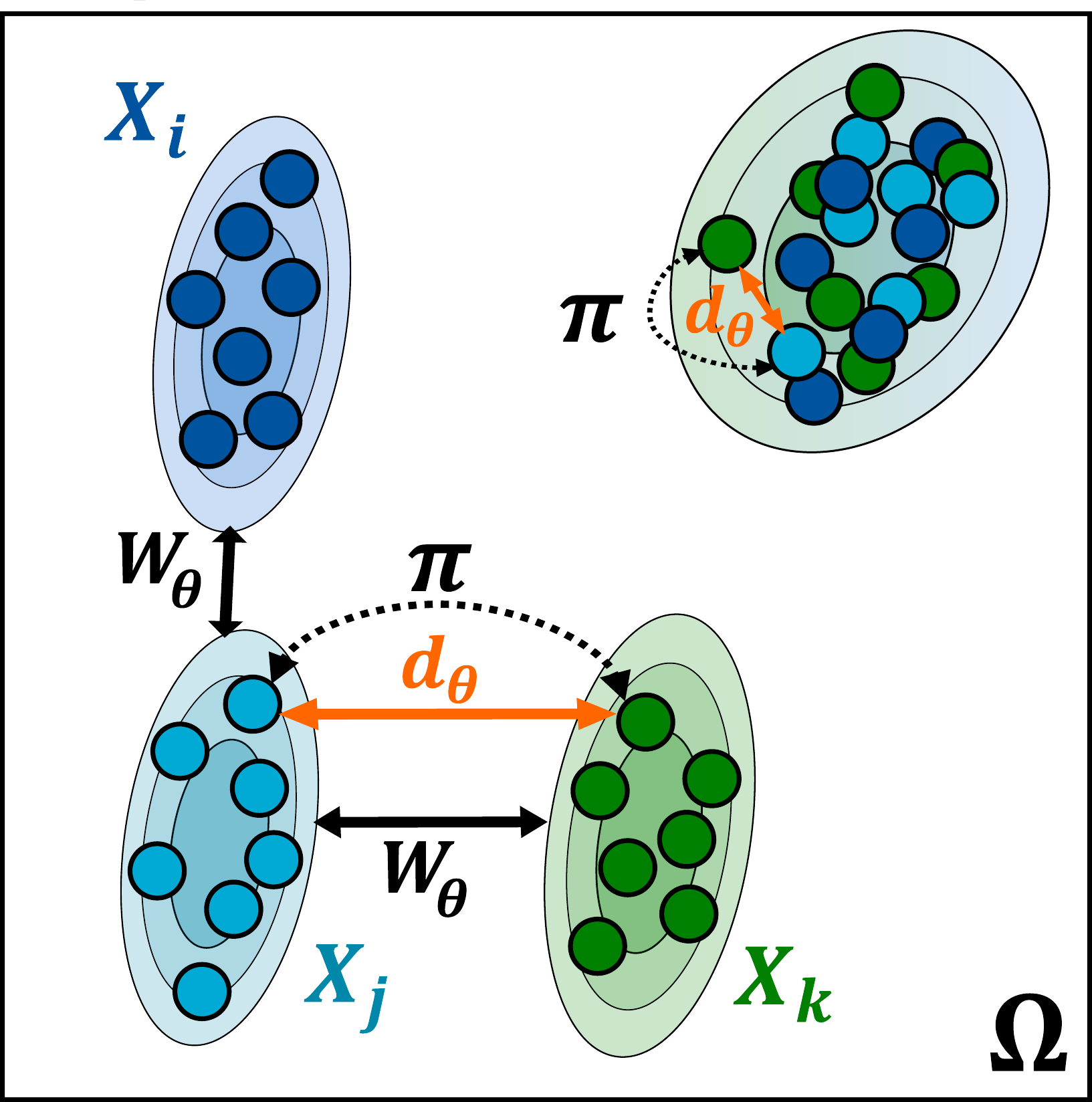} \\
        \includegraphics[trim=0.0cm 27.9cm 0.cm 0.cm, clip, width=0.49\linewidth]{figures/overview/global_triplet_ggml_overview_A.pdf} & \includegraphics[trim=0.0cm 27.9cm 0.cm 0.cm, clip, width=0.49\linewidth]{figures/overview/global_triplet_ggml_overview_B.pdf} \\
    \end{tabular}
    \caption{(a) Global ML aims to learn from relative relationships between points from different classes, without considering multimodal-distributions in a class.  
    (b) Global Ground ML learns from relative relationships between distributions from different classes, enabling to learn a metric that: globally captures relationships in different modes and, used as a ground metric, improves Wasserstein distances.}
    \label{fig:enter-label}
\end{figure}

\paragraph{Global Metric Learning.}
Global Metric Learning is a framework to learn a parameterized distance metric $d_\theta(\bm{x}_i,\bm{x}_j)$ for some ground truth distances $d^*(\bm{x}_i,\bm{x}_j)$. The Mahalanobis distance is a popular choice for such a metric $d_\theta$, where the parameters are $\theta = \bm{M}$ or equivalently $\bm{W}$. 
As ground truth distances are rarely available, most Global ML methods rely on the availability of a set of pairwise similar points $\mathcal{P}_{\approx}$ and pairwise dissimilar points $\mathcal{P}_{\not\approx}$ to learn the metric.
Global ML then typically amounts to solving an optimization problem of the form (see also \autoref{sec:related_work}):
\begin{gather}
\min\limits_\theta  \sum\limits_{(i,j)\in\mathcal{P}_\approx} d_\theta(\bm{x}_i,\bm{x}_j) \label{eq:globalml}\\
s.t. \: \forall (\bm x_j,\bm x_k)\in\mathcal{P}_{\not\approx}: \: d_\theta(\bm{x}_j,\bm{x}_k)  \geq 1  \nonumber
\end{gather}
where the learned metric $d_\theta$ minimizes the distance between pairs of similar points $(\bm x_i,\bm x_j)\in\mathcal{P}_\approx$ while satisfying a distance margin between dissimilar points $(\bm x_j,\bm x_k)\in\mathcal{P}_{\not\approx}$. 

\paragraph{Triplet Learning.}
\label{par:triplet_loss}
Triplet learning (also known in terms of triplet loss in deep learning)
was originally introduced in the context of (global) metric learning under the term Relative Comparisons \cite{schultz2003learning}. The idea is to use triplets $\mathcal{T} = \left\{(i,j,k) \middle| \:   d^*(\bm x_j,\bm x_k) - d^*(\bm x_i,\bm x_j) \geq 1 \right\} $ that capture the relative relationships between distributions. 
More specifically, $\mathcal{T}$ contains exactly all triplets $(i,j,k)$ where the distance between $\bm x_i$ and $\bm x_j$ is smaller than between $\bm x_j$ and $\bm x_k$ by at least 1. 
The corresponding metric learning (optimization) problem can then be formulated as:
\begin{gather}
\min\limits_\theta \: \frac{1}{2} \, \lVert \, \theta \, \rVert^2\\
s.t. \: \forall (i,j,k)\in\mathcal{T}: \: d_\theta(\bm{x}_j,\bm{x}_k) - d_\theta(\bm{x}_i,\bm{x}_j) \geq 1 \nonumber
\label{eq:triplet}
\end{gather}
which aims at finding a regularized $\theta$ that satisfies the constraints on the relative relationships. 

\paragraph{Wasserstein distance (EMD).}
Given two (empirical) distributions $X,Y$, the Wasserstein distance, also called Earth Movers Distance (EMD), is given as:
\begin{align}
W(X,Y) = \min\limits_{\pi} \sum_{x,y}  d(x,y) \bm{\pi}_{x,y} 
\end{align} 
where $\bm{\pi}$ is an admissible coupling (or transport plan) which maps points $x\sim X$ to $y\sim Y$ \cite{peyre2019computational}. 
Here $d$ is typically an underlying (a priori) defined ground metric such as the Euclidean, Manhattan and Cosine distance.

\paragraph{Ground Metric Learning.}
\label{par:gml}
We consider the notion of Ground Metric Learning as learning a distance function $d_\theta(x,y): \Omega \times \Omega \rightarrow \mathbb{R}_{\geq 0}$. For a discussion of different approaches please refer to \autoref{sec:related_work}.

Let $d_\theta(x,y)$ be a parameterized metric between two data points $x,y$ that is partially differentiable with respect to $\theta$. Using $d_\theta$ as the ground metric yields a parameterized Wasserstein distance:
\begin{align}
W_\theta(X,Y) = \min\limits_{\pi} \sum_{x,y}  d_\theta(x,y) \bm{\pi}_{x,y} \label{eq:pair_gml}
\end{align} 
The ground metric learning problem now is to learn $d_\theta$ such that $W_\theta(X,Y)$ approximates a given ground truth distance $W^*(X,Y)$. 
Specifically we want to find the parameters for the ground metric $d_\theta$ that minimizes the prediction error $|W_\theta(X,Y) - W^*(X,Y)|$. 
We observe that even for two distributions and ground truth distances available, learning the ground metric now becomes a nested optimization problem. 
The problem is biconvex with respect to the optimal transport plan $\pi$ between $X,Y$ and the parameters $\theta$ of the underlying ground metric $d_\theta$. 

\section{GLOBAL GROUND METRIC LEARNING}
\label{sec:ggml}
Let $X_1,...,X_n$ be probability distributions over the space $\Omega$, each labeled with a corresponding class $c_1,...,c_n\in\{1,...,k\}$. The general aim is to learn a Wasserstein distance between these distributions that separates the classes. As described in \autoref{eq:pair_gml}, the underlying ground metric $d_\theta$ between points $\bm{x}_i \sim  X_i,\bm{x}_j \sim  X_j$ can be adjusted. 
Hence, we learn an underlying ground metric that assigns distances between individual data points like a global metric, but only requires distance information on the distribution level for training like a ground metric.

To learn $d_\theta$ with Ground ML using a triplet loss, we would like to construct a set of triplets $\mathcal{T} = \left\{(i,j,k) \middle| \:  W^*(X_j,X_k) - W^*(X_i,X_j) \geq \alpha  \right\} $ containing relative distance relationships. 
However, as typically no distance information is available but only class labels, we approximate $\mathcal{T}$ by $\tilde{\mathcal{T}}= \left\{(i,j,k) \middle|   c_i = c_j \land c_j \neq c_k  \right\}$. 
This set $\tilde{\mathcal{T}}$ contains exactly the triplets $(i,j,k)$ where the empirical distributions $X_i,X_j$ belong to the same class while $X_j,X_k$ do not. Intuitively, we learn a ground metric such that distributions from the same class are closer while distributions from different classes are further apart.

\begin{assumption} Distributions $X_i,X_j$ from the same class $c_i = c_j$ are closer than distributions $X_j,X_k$ from different classes $c_j \neq c_k$. Specifically, there exists a margin $\alpha\in\mathbb{R}_{>0}$ between the distances $W^*(X_i,X_j)$ of distributions from the same class and the distance $W^*(X_j,X_k)$ of distributions from different classes.
\label{ass:class_cluster}
\end{assumption}
Under assumption \ref{ass:class_cluster}, it holds that $\tilde{\mathcal{T}} \subseteq \mathcal{T}$.
However, as the size of $\tilde{\mathcal{T}}$ still scales with $\mathcal{O}(n^3)$, we introduce a neighbor parameter $t$ that controls how many neighbors of distribution $j$ are considered to form triplets $(i,j,k)$. See \autoref{sec:related_work} for related approaches.
For each $j$, we take the Cartesian product of $t$ neighbors from the same class, indexed by $i$, and $t$ neighbors from the $C-1$ other classes, indexed by $k$. 
This significantly improves the scalability as the number of classes $C$ typically does not grow with the number of data points $n$. Under this assumption, the set of triplets $\tilde{\mathcal{T}}_t$ on which we optimize the ground metric has size $|\tilde{\mathcal{T}}_t| = n t^2  (C-1)$, i.e., scales linearly in $\mathcal{O}(n)$. 

We now aim at learning a metric $d_\theta$ such that for all triplets $(i,j,k)\in\tilde{\mathcal{T}}_t$ it holds that $ W_\theta(X_j,X_k) - W_\theta(X_i,X_j) \geq \alpha $, or equivalently $W_\theta(X_i,X_j) - W_\theta(X_j,X_k) + \alpha \leq 0$.
We write $W_\theta \approx_\alpha W^*$ if this condition is fulfilled. 
To allow for errors, we relax this hard constraint and instead formulate a minimization problem over all triplets. 
The corresponding (unregularized) loss function to learn global ground metrics from distributions with class labels is defined as:
\vspace{2mm}
\begin{align}
\mathcal{L}_\alpha(\theta,X,\tilde{\mathcal{T}_t}) &=  \sum\limits_{t\in\tilde{\mathcal{T}_t} } \mathcal{L}_\alpha(\theta,X,t)
\label{eq:loss} 
\end{align} 
\vspace{-5mm}
\begin{multline}
\text{where} \: \mathcal{L}_\alpha(\theta,X,(i,j,k)) = \\
\max \left( W_\theta(X_i,X_j) - W_\theta(X_j,X_k) + \alpha, 0 \right) 
\label{eq:partial_loss}
\vspace{3mm}
\end{multline}

To balance the influence of the unbounded $- W_\theta(X_j,X_k)$ term on the objective function, we use a ReLU activation function to bound the margin of each triplet relationship in \autoref{eq:partial_loss}.
Intuitively, the loss corresponds to the sum of errors of all triplet relationships separated by a margin less than $\alpha$. 

\begin{thm} 
The GGML loss is 0 if, and only if, the global ground metric $d_\theta$ in $W_\theta$ approximates the ground truth distances $W^*$ with margin $\alpha$. 
$$\mathcal{L}_\alpha(\theta,X,\tilde{\mathcal{T}_t})\!=\!0 \: \text{ iff } \:  W_\theta \approx_\alpha W^*$$
\end{thm} 
This theorem states the desirable property that the minimal loss is only 0 if the ground metric $d_\theta$ in $W_\theta$ separates relative relationships between classes by at least margin $\alpha$.

If there exist $\theta' \neq \theta$ such that $d_{\theta} = d_{\theta'}$, the optimization of the loss \eqref{eq:loss} can become ill-posed. This is certainly the case for the Mahalanobis distance (\autoref{eq:mahala}) as uniqueness of $\bm W$ is not guaranteed in the matrix decomposition. To facilitate unique optimal solutions, we thus introduce the regularized loss function:
\begin{align}
\mathcal{L}_{\alpha,\lambda}(\theta,X,\tilde{\mathcal{T}_t}) &=  \sum\limits_{t\in\tilde{\mathcal{T}_t} } \mathcal{L}_\alpha(\theta,X,t) + \lambda R(\theta) \label{eq:loss1}\end{align} 
where $R(\theta)$ acts as a (differentiable) regularization term on $\theta$. Parameter $\lambda\in \mathbb{R}_{\geq 0}$ controls the regularization strength. In our numerical evaluation on scRNA data, we regularize with the Frobenius norm to avoid overfitting and improve generalizability. To increase the sparsity of the learned $\theta$, an L1 norm can be used instead, which we demonstrate on synthetic data. For different purposes, other differentiable regularization terms may be used.

Learning $\theta$ by optimizing $\mathcal{L}(\theta,X,\tilde{\mathcal{T}_t})$ corresponds to a finite sum problem over $\mathcal{T}_t$. By construction $d_\theta$, and hence $W_\theta$, is partially differentiable with respect to $\theta$ which enables optimization using gradient descent. 
Despite our approach facilitating a linear complexity of the triplet set $\mathcal{T}_t$, this optimization can become difficult to solve for large datasets. 
To address this, we perform Stochastic Gradient Descent (SGD) over subsets of $\mathcal{T}_t$ as mini batches. Unlike previous SGD approaches in Ground ML~\cite{wang2012supervised}, we do not alternate between optimizing the transport plan and the ground metric. Instead, we directly compute the full gradient over the nested optimization problem which is guaranteed to exist. 

\subsection{Low Rank Mahalanobis Distance as Global Ground Metric}
\label{sub:low_rank_mahala}
We illustrate the applicability of our approach using a Mahalanobis distance (\autoref{eq:mahala}). 
More specifically, we aim to learn a Mahalanobis distance as our ground metric $d_\theta$ with $\theta = \bm{W}$.
Interestingly, when using the Mahalanobis distance as a Global Ground Metric in our framework, we can give upper bounds on the partial triplet loss and the total loss.
\begin{thm}[Triplet Loss Bound]
For all distributions $X$ and triplets $(i,j,k)\in\tilde{\mathcal{T}}_t$, there exist a $\theta$ such that $\mathcal{L}(\theta,X,(i,j,k))$ 
is at most $\alpha$. 

For all distributions $X$ and sets of triplets $\tilde{\mathcal{T}}_t$, the minimal loss $\mathcal{L}(\theta,X,\tilde{\mathcal{T}_t})$ is at most $\alpha  |\tilde{\mathcal{T}_t}|$. \label{lem:loss_bound} 
\end{thm} 

The number of parameters of a Mahalanobis distance scales quadratically with feature dimensions which prohibits its application to large datasets. 
To facilitate efficient computations on large datasets, we propose an approximation of the Mahalanobis matrix $\bm{M}$ by some low rank matrix factorization such that $\bm{\widetilde{W}}^T \bm{\widetilde{W}} \approx \bm{W}^T \bm{W} = \bm{M}$, where $\bm{\widetilde{W}}$ is of rank $k \ll n$.
For low rank, symmetric $\bm{M}$ of rank $k$, this approximation is of course exact. In practice, we observe that the approximation is accurate for low rank $\bm{M}$ and becomes less accurate for higher rank.

\begin{figure*}[t!]
\centering
    \vspace*{-0.5cm}
    \resizebox{0.85\linewidth}{!}{
    \begin{tabular}{@{}rcccc@{}}
    
      \makebox[0pt]{\raisebox{60pt}{\rotatebox[origin=c]{90}{\textbf{Sampled Points}}}}
    & \multicolumn{4}{c}{
    \subfloat[Multimodal distributions with class labels]{\includegraphics[trim=.2cm 0.3cm .cm 0.cm, clip, width=0.38 \textwidth]{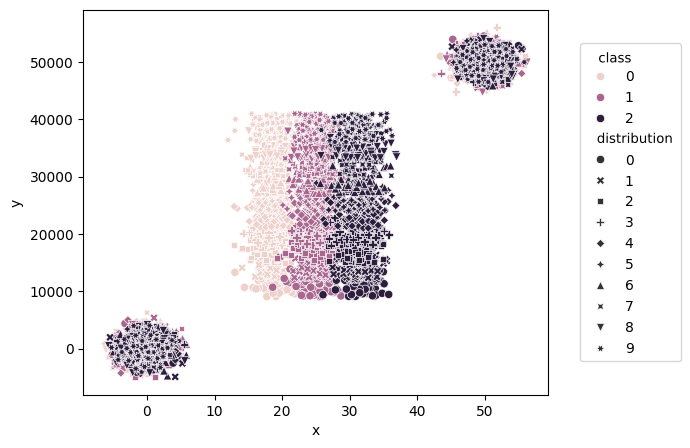}}\hspace{0.5cm}%
    \subfloat[Transformation of points]{\includegraphics[trim=0.0cm 0.3cm 2.cm 0.7cm, clip, width=0.31 \textwidth]{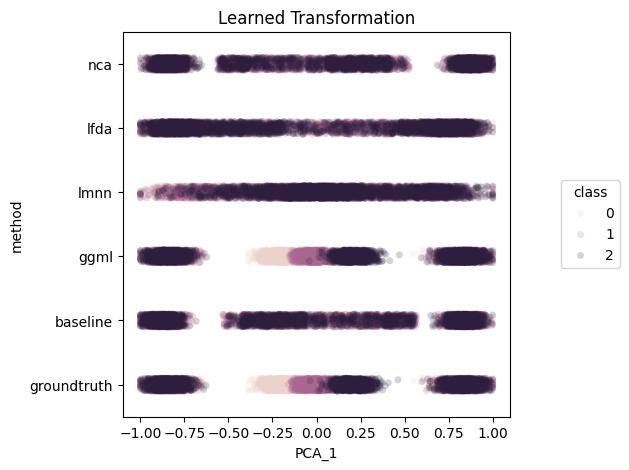}}\hspace{0.5cm}%
    \subfloat[Mahalanobis matrix $\bm{M}$]{\includegraphics[trim=0cm  0.15cm 0cm  0.25cm, clip, width=0.25 \textwidth]{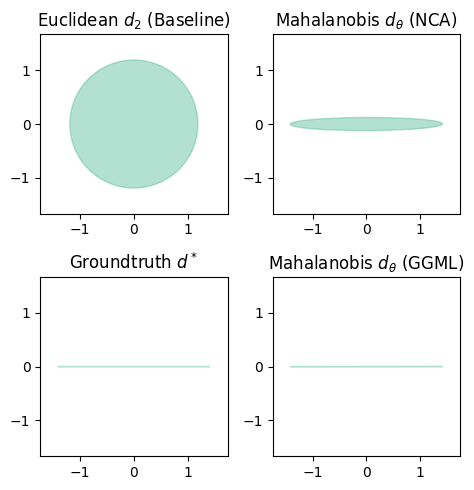}}} \\[-1ex]
    \makebox[0pt]{\raisebox{47pt}{\rotatebox[origin=c]{90}{\textbf{Distributions}}}} & 
    \subfloat{\includegraphics[trim=0.0cm 0.0cm 0.cm 0.cm, clip, width=0.22 \textwidth]{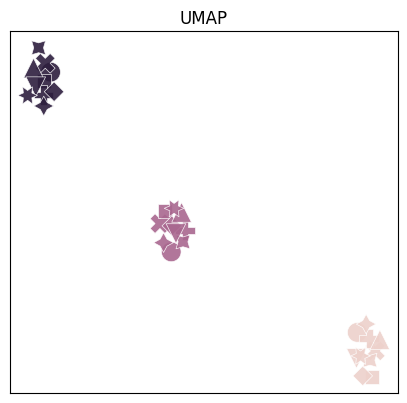}} &
    \subfloat{\includegraphics[trim=0.25cm 0.0cm 0.25cm 0.cm, clip, width=0.23 \textwidth]{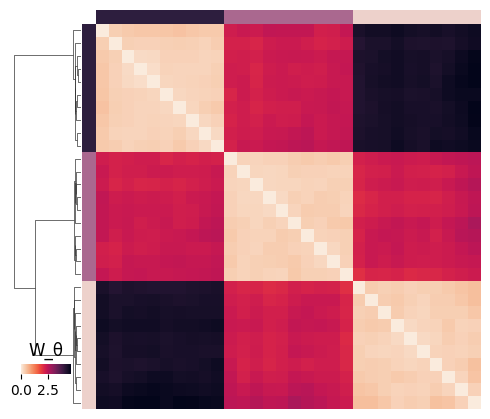}}\hspace{0.8cm} &
    \subfloat{\includegraphics[trim=0.0cm 0.0cm 0.cm 0.cm, clip, width=0.22 \textwidth]{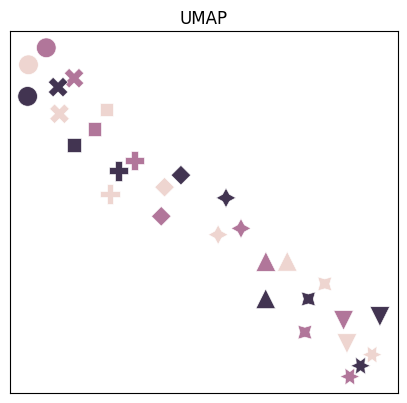}} &
    \subfloat{\includegraphics[trim=0.25cm 0.0cm 0.25cm 0.cm, clip, width=0.23 \textwidth]{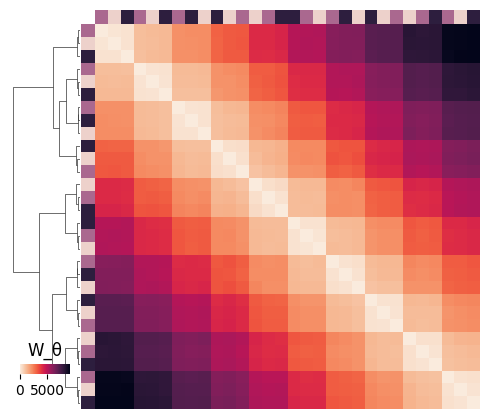}}\\
    &  Embedding &Clustermap &  Embedding &Clustermap \\
    & \multicolumn{2}{c}{(d) Optimal transport with GGML} &  \multicolumn{2}{c}{(e) Optimal transport with Euclidean ground metric} \\
    \end{tabular}}
    \vspace*{-0.1cm}
    \caption{(a) Sampled points from the two-dimensional synthetic dataset with noise on the y-axis. (b) Transformations learned by competing methods show that only GGML recovers the signal on the x-axis that differentiates distributions. (c) The corresponding Mahalanobis matrix which is represented as a normalized ellipse shows the learned relative scale of the axes. (d) Optimal transport of the multimodal distributions with GGML separates the classes, which (e) a Euclidean ground metric does not.}    
    \label{fig:synthetic_ggml}
    \vspace*{-0.2cm}
\end{figure*}

Intuitively, the existence of a low rank Mahalanobis (covariance) matrix $\bm{M}$ corresponds to the data being supported in some lower-dimensional subspace. 
Our low rank method enables us to efficiently learn such subspaces that capture class relations. Other approaches to address high dimensionality (e.g., PCA) reduce the feature space to some lower-dimensional subspace that does not necessarily capture such class relationships. 
More specifically, methods like PCA aim to capture all aspects of the data which might include a lot of variance shared between classes (i.e. unrelated signals or noise).
Using GGML to learn a low rank Mahalanobis distance thus enables us to differentiate class-related variations in the data from noise and class-unrelated signals.

\section{APPLICATIONS TO SYNTHETIC AND REAL-WORLD SCRNA-SEQ DATA}
\label{sec:applications}
To evaluate our novel framework, we perform several tasks on synthetic and real-world scRNA-seq datasets.
In this section, we describe the setup and evaluation of different tasks using distances $d$ resp. $W$ learned by various competing methods, introduced in \autoref{sec:related_work}. 
\paragraph{Classification} serves as a benchmark to evaluate how effectively the learned distances from various metric learning methods reflect the relationships between classes.
We perform classification on both the cell- and patient-level using a weighted $k$-NN \cite{guo2003knn}. This classifier assigns class labels to data points based on the weighted majority class of its $k$-closest neighbors under $d$ resp. $W$. To evaluate generalizability, we measure the prediction accuracy over 10 test-train splits with respective half of the data points. We are withholding 20\% of data points in each split for the hyperparameter tuning in \autoref{sec:hyper}. Test-train splits are done under a group shuffle split such that cells from a specific patient only occur in a given train or test set.
\paragraph{Clustering} of data points and distributions is performed with agglomerative hierarchical clustering \cite{murtagh2012algorithms}. This method builds hierarchical clusters by iteratively merging clusters based on the learned distances. It is evaluated using common clustering metrics, see appendix \ref{ssec:clust}
\paragraph{Embedding} into a low dimensional space that approximates the learned $d$ resp. $W$. Embeddings on the distribution-level show the relationships between patients under a Wasserstein distance $W$ using $d$ as a ground metric. The cell-level shows the relationships between cells with $d$ as a global metric. We use UMAP to compute 2D embeddings of the learned distances for visualization purposes \cite{mcinnes2018umap}.
\paragraph{Feature Importance}\label{par:feat_imp} is derived by interpreting rows of $\widetilde{\bm{W}}$ as distinct class-related processes. To get gene importance values over all class-related processes, we reconstruct the full Mahalanobis matrix as $\bm{M}\approx \widetilde{\bm{W}}^T\widetilde{\bm{W}}$. Diagonal entries correspond to the features relative importance in differentiating the classes. Off-diagonal entries correspond to learned relative importance of interactions (i.e. correlations). For scRNA data, this feature importance can be interpreted as the relative importance of expressed genes in explaining disease states. Here, we qualitatively describe the identified genes in the literature for the respective disease. An exemplary gene enrichment analysis is provided in the appendix \ref{ssec:gene_enrich}.
\begin{figure*}
  \vspace*{-0.1cm}
  \resizebox{0.98\linewidth}{!}{
  \centering
  \setlength{\tabcolsep}{1pt}
  \renewcommand{\arraystretch}{0.1}
  \begin{tabular}{lcccccccc}
  & Patient-level Clustering & \multicolumn{2}{c}{Patient-level Embedding} &  \multicolumn{2}{c}{Cell-level Embedding \textbf{GGML} $d_\theta$} & Feature Importance \\
  & (a) \textbf{GGML} $W_\theta$ & (b) \textbf{GGML} $W_\theta$ &  (c) Baseline $W_2$ & (d) Disease & (e) Cell type & (f) Genes in $\theta$ \vspace{-2mm}\\
  \multirow{1}{*}[80pt]{\rotatebox[origin=B]{90}{\textbf{Breast cancer}}} &%
  \subfloat{\includegraphics[width=0.23 \textwidth]{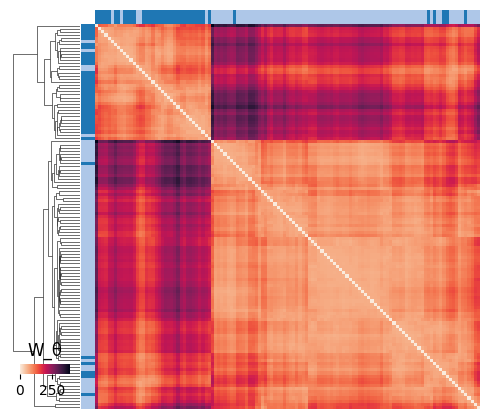}}     &
  \subfloat{\includegraphics[trim=.0cm 0.0cm .cm 0.cm, clip, width=0.2 \textwidth]{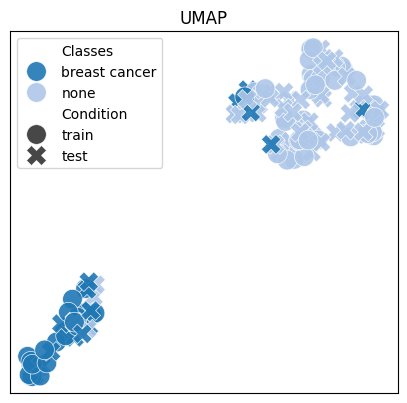}}&
  \subfloat{\includegraphics[trim=.0cm 0.0cm .cm 0.cm, clip, width=0.2 \textwidth]{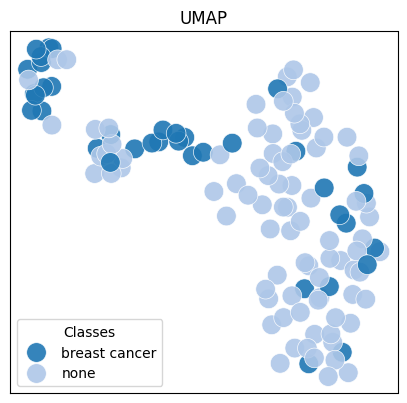}}&
  \subfloat{\includegraphics[trim=.0cm 0.0cm .cm 0.cm, clip, width=0.2 \textwidth]{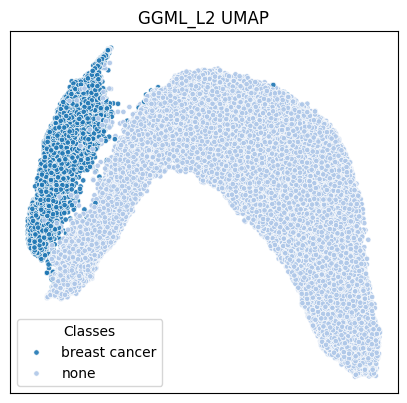}} &
  \subfloat{\includegraphics[trim=.0cm 0.0cm .cm 0.cm, clip, width=0.2 \textwidth]{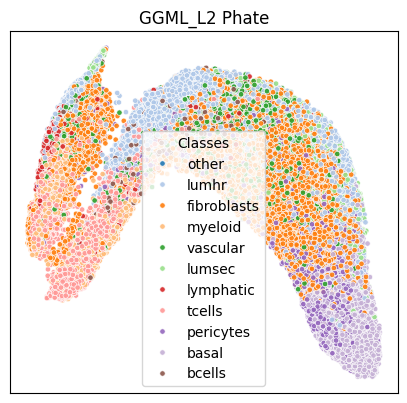}}&
  \subfloat{\includegraphics[trim=.0cm 0.cm 0.cm 0.cm, clip, width=0.165 \textwidth]{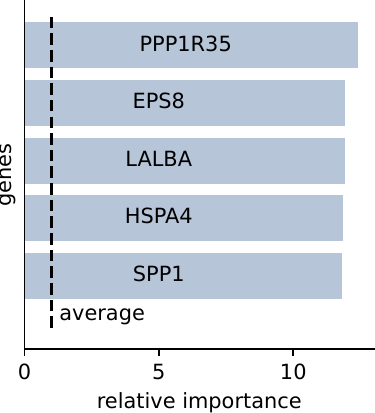}} \vspace{-3mm}\\
  \multirow{1}{*}[85pt]{\rotatebox[origin=t]{90}{\textbf{Kidney disease}}} &
  \subfloat{\includegraphics[width=0.23 \textwidth]{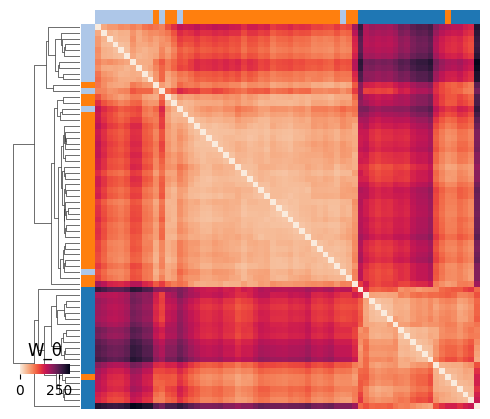}} &
  \subfloat{\includegraphics[trim=.0cm 0.0cm .cm 0.cm, clip, width=0.2 \textwidth]{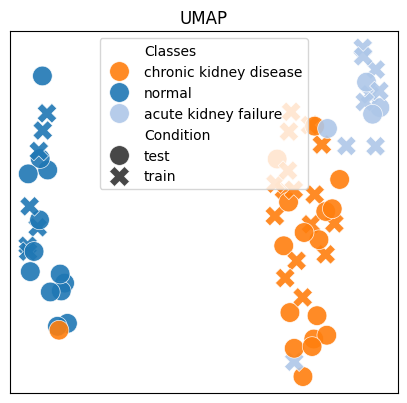}}& 
  \subfloat{\includegraphics[trim=.0cm 0.0cm .cm 0.cm, clip, width=0.2 \textwidth]{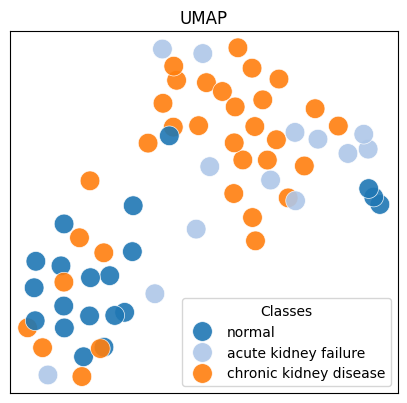}}& 
  \subfloat{\includegraphics[trim=.0cm 0.0cm .cm 0.cm, clip, width=0.2 \textwidth]{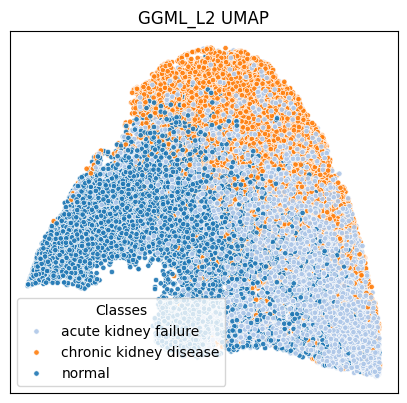}}&  
  \subfloat{\includegraphics[trim=.0cm 0.0cm 0.cm 0.cm, clip, width=0.2 \textwidth]{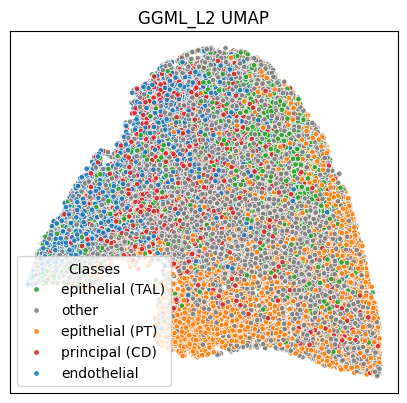}}&
  \subfloat{\includegraphics[trim=.0cm 0.cm 0.cm 0.cm, clip, width=0.165\textwidth]{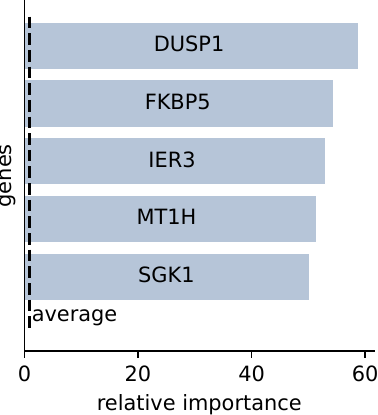}} \vspace{-3mm}\\
  \multirow{1}{*}[80pt]{\rotatebox[origin=c]{90}{\textbf{Myocard. inf.}}} &%
  \subfloat{\includegraphics[width=0.23 \textwidth]{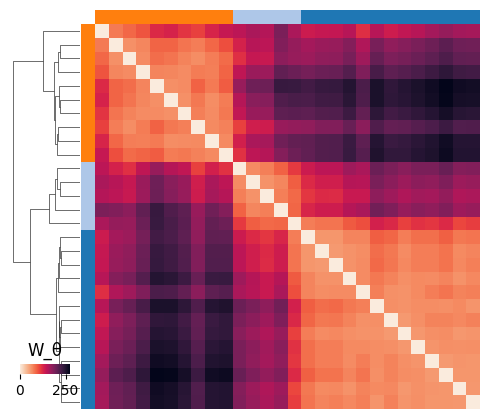}} &
  \subfloat{\includegraphics[trim=.0cm 0.0cm .0cm 0.cm, clip, width=0.2 \textwidth]{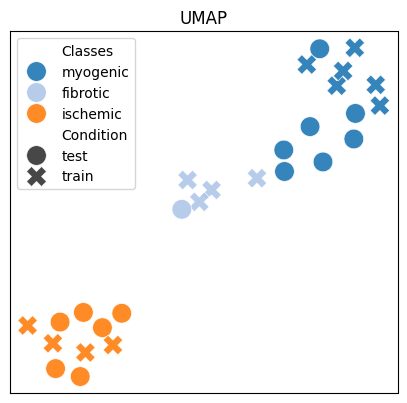}}&
  \subfloat{\includegraphics[trim=.0cm 0.0cm .0cm 0.cm, clip, width=0.2 \textwidth]{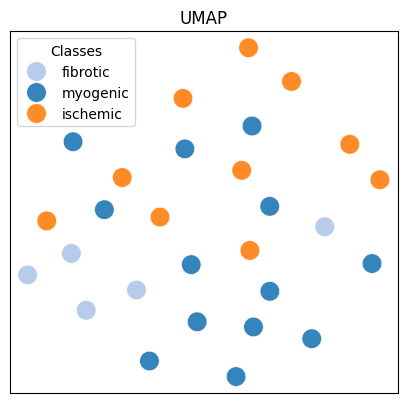}}&
  \subfloat{\includegraphics[trim=.0cm 0.0cm 0.cm 0.cm, clip, width=0.2 \textwidth]{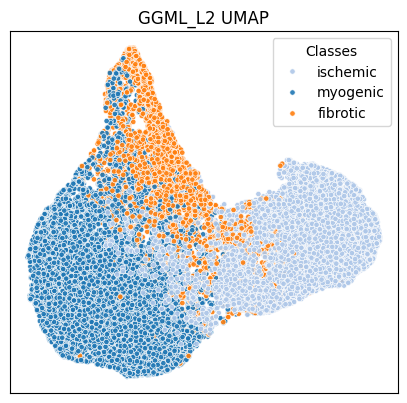}} & 
  \subfloat{\includegraphics[trim=.0cm 0.0cm 0.cm 0.cm, clip, width=0.2 \textwidth]{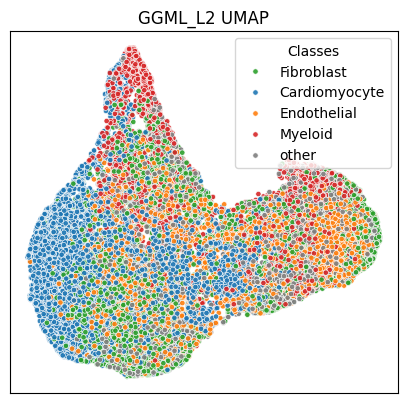}} & 
  \subfloat{\includegraphics[trim=.0cm 0.cm 0.cm 0.cm, clip, width=0.165 \textwidth]{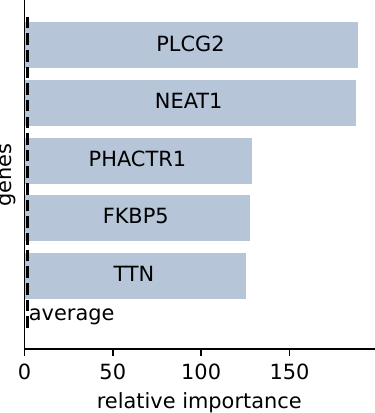}}\\
  \end{tabular}
  }
  \vspace*{-0.075cm}
  \caption{Embeddings of patients and cells for scRNA-seq data from different diseases using $d_\theta$ learned by Global Ground Metric Learning (GGML) and Euclidean $d_2$ as baseline. To highlight the capabilities of GGML to generalize to unseen data the shown plots are produced by only learning on half of the data points as indicated.
  Relative weights of $\theta$ can be directly interpreted as gene importance in distinguishing disease stages. }
  \label{fig:scRNA_results}
\end{figure*}

\subsection{Synthetic Datasets}

\label{gen_inst}
To validate our approach in a controlled setting, we generate synthetic data with reliable ground truth. Points $\bm{x_i}$ in this dataset are distributed according to three-modal distributions $X_i$ from three classes.  
As shown in \autoref{fig:synthetic_ggml}, the three classes are only distinguishable in one of the modes (center) and along one axis. The other two modes are identically distributed and correspond to some unrelated process. For ease of explanation and visual clarity, no rotations are applied to the synthetic example.
The first synthetic dataset shown in \autoref{fig:synthetic_ggml}(a), contains two dimensional data for visualization purposes. While varying scales on the axes are used to demonstrate the influence of noise in the two dimensional setting, it directly translates to problems arising from small noise levels in large dimensional data. We demonstrate this on the second synthetic dataset with 200 identically scaled dimensions. Hyperparameters are tuned with a grid search evaluated on 20\% validation splits as shown in \autoref{fig:hyper}. 

\subsection{scRNA-seq Datasets}
Single-cell RNA sequencing data contains measured gene expression vectors $\bm{x}$ of individual cells from different patients. 
Disease (i.e. class) labels $c_i$ are generally provided as pathological annotations of the respective tissue and correspond to classes on the patient-level. This setup fits naturally in our framework with sampled cells $\bm{x}$ from a multimodal patient-specific distribution of cells $X_i$ for some patient $i$.  
Similar to the synthetic data, the assumption is that disease stages are not distinguishable for cells from cell types (modes) that are not disease-relevant, presenting a realistic and challenging scenario for testing our approach.
We perform our experimental evaluation on datasets from three distinct diseases: breast cancer \cite{sikkema2023integrated}, kidney disease \cite{lake2023atlas} and myocardial infarction \cite{kuppe2022spatial}. This diverse collection of large datasets allows us to benchmark our framework across a variety of biological contexts and disease states. The datasets contain 31 to 132 tissue samples with a total of 191k to 714k cells from which we sample 1000 cells per tissue.
The datasets included 28,975 to 33,145 genes per cell. We select only those genes with above-average variance, resulting in a reduced dimensionality of 7,734 to 8,433 genes. 

\subsection{Results}
We present the results of the evaluation on the synthetic datasets in \autoref{fig:synthetic_ggml} and the scRNA datasets in \autoref{fig:scRNA_results}. The classification results are found in \autoref{tab:class}. To reproduce the presented results or use GGML on other data, the code is provided on GitHub\footnote{\href{https://www.github.com/DaminK/GlobalGround-MetricLearning}{github.com/DaminK/GlobalGround-MetricLearning}}. 

\begin{table*}[t]
\centering
\resizebox{1\linewidth}{!}{
\footnotesize
\setlength{\tabcolsep}{1pt}
\begin{tabular}{lr||cc|cc|cc|cc|cr||cc|cc|cc|cc|ccc}
\multirow{1}{*}{Method} &  &  \multicolumn{2}{c|}{$\text{Synth}_{2D}$} & \multicolumn{2}{c|}{$\text{Synth}_{200D}$} & \multicolumn{2}{c|}{Kidney} & \multicolumn{2}{c|}{Breast} & \multicolumn{2}{c||}{Myocard.} &  \multicolumn{2}{c|}{$\text{Synth}_{2D}$} & \multicolumn{2}{c|}{$\text{Synth}_{200D}$} & \multicolumn{2}{c|}{Kidney} & \multicolumn{2}{c|}{Breast} & \multicolumn{2}{c}{Myocard.} \\
\toprule
Eucl. & \multirow{8}{*}{\rotatebox[origin=c]{90}{\it{patient-level}}} & 0.24±0.08 & & 0.39±0.12 &  & 0.52±0.10 & & 0.77±0.03 & & 0.49±0.03 &  \multirow{8}{*}{\rotatebox[origin=c]{90}{\it{cell-level}}} & 0.32±0.01 &   & 0.45±0.01 &  & 0.45±0.11 & & 0.79±0.04 & & 0.48±0.10 & \\
Manh. & & 0.24±0.08 &  & 0.39±0.11 &  & 0.57±0.08 & & 0.79±0.03 & & 0.85±0.03 & & 0.33±0.01 &  & 0.41±0.01 &  & 0.48±0.07 & & 0.79±0.04 & & 0.56±0.08 \\
Cos.  & & 0.43±0.07 &  & 0.46±0.10 &  & 0.54±0.07 & & 0.79±0.03 & & 0.53±0.06 & & 0.36±0.01 &   & 0.35±0.01 &  & 0.46±0.10 & & 0.79±0.04 & & 0.53±0.12  \\
LMNN  & & 0.22±0.07 &  & 0.29±0.11 &  & \multicolumn{2}{c|}{\textit{OOM}} & \multicolumn{2}{c|}{\textit{OOM}} & \multicolumn{1}{c}{\textit{OOM}} & & 0.38±0.01 & & 0.45±0.01 &  & \multicolumn{2}{c|}{\textit{OOM}} & \multicolumn{2}{c|}{\textit{OOM}} & \multicolumn{2}{c}{\textit{OOM}} \\
LFDA  & & 0.46±0.06 & &0.47±0.09 & & 0.86±0.11 & & 0.82±0.07 & &  0.88±0.11 & & 0.40±0.01 &  & 0.37±0.01 & & 0.52±0.06 & & 0.67±0.03 & & 0.88±0.07\\
NCA   & & 0.35±0.11 &  & 0.25±0.11 & & \multicolumn{2}{c|}{\textit{OOT}} & \multicolumn{2}{c|}{\textit{OOT}} & 0.81±0.06 & & 0.37±0.00 &   & 0.46±0.02 & & \multicolumn{2}{c|}{\textit{OOT}} & \multicolumn{2}{c|}{\textit{OOT}} & 0.78±0.08 \\
ITML  & & 0.51±0.09	&  & 0.43±0.08 & & 0.55±0.10 & & 0.76±0.04 & & 0.79±0.04 & & 0.41±0.01 &   & 0.36±0.01 & & 0.37±0.06 & & 0.77±0.04 & & 0.54±0.07 & \\
GGML  & & \textbf{0.96±0.04} &  & \textbf{0.95±0.11} & & \textbf{0.94±0.02} & & \textbf{0.91±0.04} & & \textbf{0.92±0.08} &  & \textbf{0.53±0.01} &  & \textbf{0.53±0.01} & & \textbf{0.74±0.00} & & \textbf{0.81±0.03} & & \textbf{0.94±0.00} & \\
\end{tabular}}
\caption{kNN Classification accuracy on patient- and cell-level for competing metric learning methods. Accuracy is given as mean±variance over 10 test-train splits. \textit{OOM/T} stands for \textit{Out of Memory/Time}. }
\label{tab:class}
\end{table*}
\paragraph{Classification}
\autoref{tab:class} presents classification results of various methods using a kNN classifier on 10 test-train splits of the patient-level and cell-level data. For patient-level data, the GGML method consistently outperforms other methods such as Euclidean (Eucl.), Manhattan (Manh.), Cosine (Cos.), LMNN, LFDA, NCA, and ITML across different datasets. It achieves high accuracies on synthetic data (2D: 0.96±0.04, 200D: 0.95±0.11) and real-world genomics datasets from different diseases (kidney: 0.94±0.02, breast cancer: 0.91±0.04, myocard: 0.92±0.08). 
On the cell-level, GGML again shows superior performance on all datasets (synth: 0.53±0.01, kidney: 0.74±0.00, breast cancer: 0.81±0.03, myocard: 0.94±0.00). As expected, differentiating disease states at the cell-level is a more challenging task as not all cells are involved or affected by the disease. 
We indicate that certain methods were unable to process large-scale scRNA-seq data within the available memory limits (1TB) or computational time constraints (64 cores, 8 hours). This underscores the need for efficient algorithms capable of handling high-dimensional biological data.

\paragraph{Clustering} of patients using the learned $W$ from GGML are shown for the 2D synthetic dataset in \autoref{fig:synthetic_ggml}(e), and the scRNA datasets in \autoref{fig:scRNA_results}(a). 
The shown cluster maps clearly differentiate disease states at the patient level for kidney, breast cancer, and myocardial infarction. Notably, the clusters meaningfully capture all patients, even though GGML was only trained on half of the patients as indicated in \autoref{fig:scRNA_results}(b).
Similar to classification, GGML demonstrates superior performance compared to other methods in these clustering tasks based on standard clustering metrics.
Detailed results on the clustering performance can be found in the appendix \autoref{ssec:clust}. 

\begin{figure}[b!]
    \footnotesize
    \centering
    \setlength{\tabcolsep}{1pt}
    \renewcommand{\arraystretch}{0}
    \begin{tabular}{lccc}
    & (a) \textbf{GGML} & (b) Average & (c) Euclidean  \vspace{-2mm}\\
    \multirow{1}{*}[50pt]{\rotatebox[origin=c]{90}{\textbf{Disease}}} & 
    \subfloat{\includegraphics[trim=.0cm 0.1cm .0cm 0.66cm, clip,width=0.3 \columnwidth]{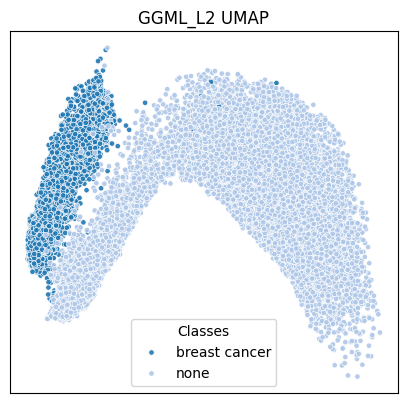}}&
    \subfloat{\includegraphics[trim=.0cm 0.1cm .0cm 0.66cm, clip,width=0.3 \columnwidth]{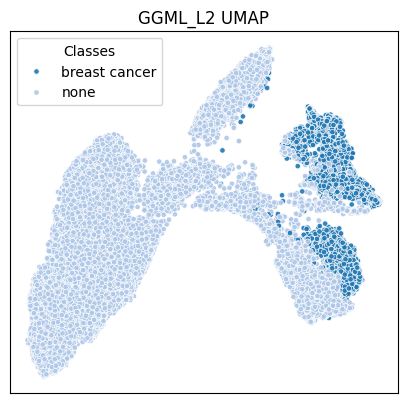}}&
    \subfloat{\includegraphics[trim=.0cm 0.1cm .0cm 0.65cm, clip,width=0.3 \columnwidth]{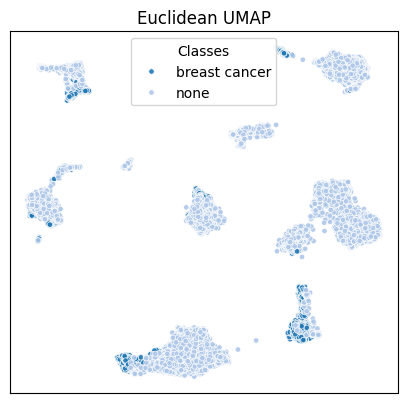}} \vspace{-2mm}\\
    \multirow{1}{*}[53pt]{\rotatebox[origin=c]{90}{\textbf{Cell type}}} &
    \subfloat{\includegraphics[trim=.0cm 0.1cm .0cm 0.66cm, clip,width=0.3 \columnwidth]{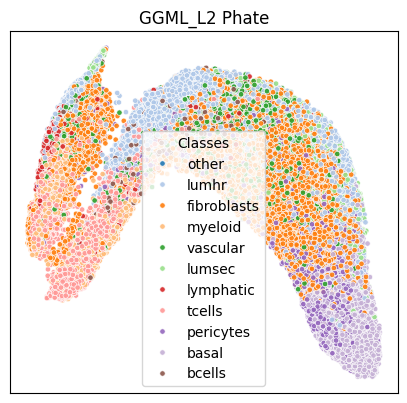}}&
    \subfloat{\includegraphics[trim=.0cm 0.1cm 11.15cm 0.66cm, clip,width=0.3 \columnwidth]{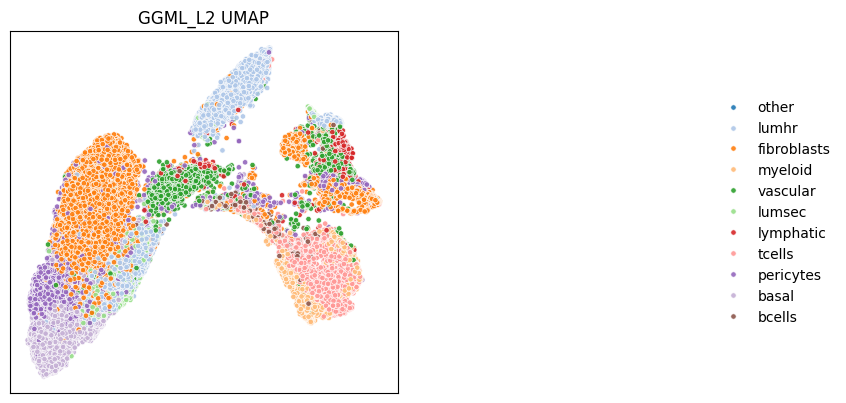}}&
    \subfloat{\includegraphics[trim=.0cm 0.1cm 11.15cm 0.65cm, clip,width=0.3 \columnwidth]{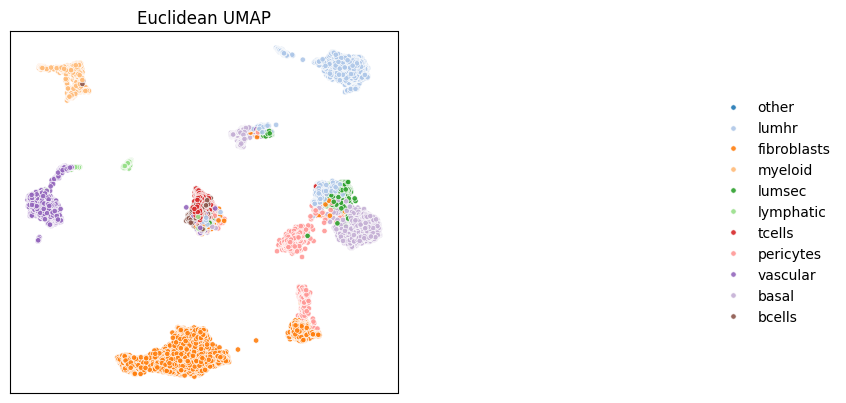}}\\
    \end{tabular}
    \caption{GGML(a) learns a Mahalanobis distance that differentiates cells by disease state. Cell types are defined by the Euclidean distance(c) which does not capture disease states. The average of both distances(b) differentiates cells by disease state and type.}
    \label{fig:average_results}
\end{figure}    

\paragraph{Embedding} of the learned Wasserstein distances between the distributions (patients) shows well separated classes in the synthetic (\autoref{fig:synthetic_ggml}d) and scRNA data (\autoref{fig:scRNA_results}b). 
To highlight the capabilities of GGML to generalize to unseen data, shown embeddings are produced by only learning on half of the data points as indicated.
For the baseline of using the euclidean $d_2$ as a predefined ground metric, the embeddings (\autoref{fig:synthetic_ggml}f,\ref{fig:scRNA_results}c) show no clear separations of classes or disease states.
Embedding the sampled points shows that GGML learns to differentiate the center mode in the synthetic data (\autoref{fig:synthetic_ggml}b) unlike other competing methods. In the synthetic data, GGML is the only method that captures the classes-related variances along a single axis shown by the first principal components of learned spaces from different methods. In the cell-level of the scRNA data, we only show embeddings of cells that are classified with 90\% accuracy over the train-test splits as a proxy for being involved in the disease progression. Baseline embeddings of all cells, including competing methods, can be found in the appendix \ref{ssec:emb}.

\vspace{-0.2cm}
\paragraph{Relating Disease States and Cell Types}
\label{sub:bary_mahala}
Analyzing scRNA data is commonly done by clustering cells into distinct cell types that are assumed to serve different functions in the tissue. These clusterings often use the Euclidean distance over all genes or their principal components as measure of cell similarity. Note that the Euclidean distance is a Mahalanobis distance with identity matrix.
To relate different Mahalanobis distances over the same cells, we aim at computing averages as weighted linear combinations of the cell distances. Specifically, computing such averages enables us to fully interpolate between the GGML Mahalanobis distance that captures disease states and the Euclidean distance that captures cell types. \autoref{fig:average_results}(b) shows that the average distance captures properties from both distances differentiating the disease state and cell type. It shows two large clusters of different cell types with disease states associated to breast cancer. This highlights its capabilities as a tool for researchers in computational biology where relating novel findings such as disease states to cell types can be crucial in understanding disease progressions.

\paragraph{Feature Importance}
For the low dimensional synthetic data, we can represent the learned Mahalanobis (covariance) matrix of different methods as 2D ellipses \autoref{fig:synthetic_ggml}(c). GGML recovers the ground truth where only variance across the first axis explains the class differences on a distribution-level.
For scRNA data, we show the five most important genes in differentiating each disease in \autoref{fig:scRNA_results}(f). The identified genes contain known markers for the respective diseases or are associated with related biological processes. 
In the disease progression of breast cancer, genes PPP1R35 and HSPA4 are associated with lymph node metastases \cite{mamoorppp1r35, gu2019tumor}, while EPS8 regulates cell migration and proliferation \cite{chen2015eps8}.  
SPP1 has a role in angiogenesis and fibroblast activation in the tumor micro-environment \cite{butti2021tumor}.
In the kidney dataset, the identified genes impact both chronic kidney disease (CKD) and acute kidney injury (AKI). 
DUSP1 and FKBP5 were found to protect against AKI and CKD caused by ischemia \cite{shi2023dusp1} respective inflammatory responses \cite{xu2022microrna}. IER3 affects the same signalling pathways related to inflammation \cite{arlt2011role}. 
Elevated levels of SGK1 expression have been demonstrated to promote hypertrophy and fibrosis, leading to kidney damage in a mouse model \cite{sierra2021increased}.
In the disease progression of myocardial infarction, PHACTR1 is related to affecting arterial compliance which can be understood as flexibility of the vascular system \cite{wood2023phactr1}. 
NEAT1 is related to early onset of myocardial infarction involved in multiple pathways related to immune and inflammatory responses \cite{gast2019long}, with PLCG2 being related to similar processes causing immune dysregulation \cite{welzel2022variant}.
FKBP5, a gene already found to be relevant in kidney disease, is also associated with increasing risk for myocardial infarction contributing to the same signal pathway \cite{zannas2019epigenetic}. 
These findings, which align with existing literature across all considered diseases, highlight the interpretability of the learned weights as indicators of gene importance in disease-related processes. An exemplary gene enrichment analysis using the learned weights can be found in \autoref{fig:biological_process}.

\begin{figure}
    \footnotesize
    \centering
    \setlength{\tabcolsep}{1pt}
    \renewcommand{\arraystretch}{0}
    \begin{tabular}{lcc}
    & (a) Synth. (200D)& (b) Myocard. (8744D) \vspace{-2mm}\\
    \multirow{1}{*}[68pt]{\rotatebox[origin=c]{90}{\textbf{Patient-level}}} & 
    \subfloat{\includegraphics[trim=.0cm 0.cm .0cm 1.cm, clip,width=0.45 \columnwidth]{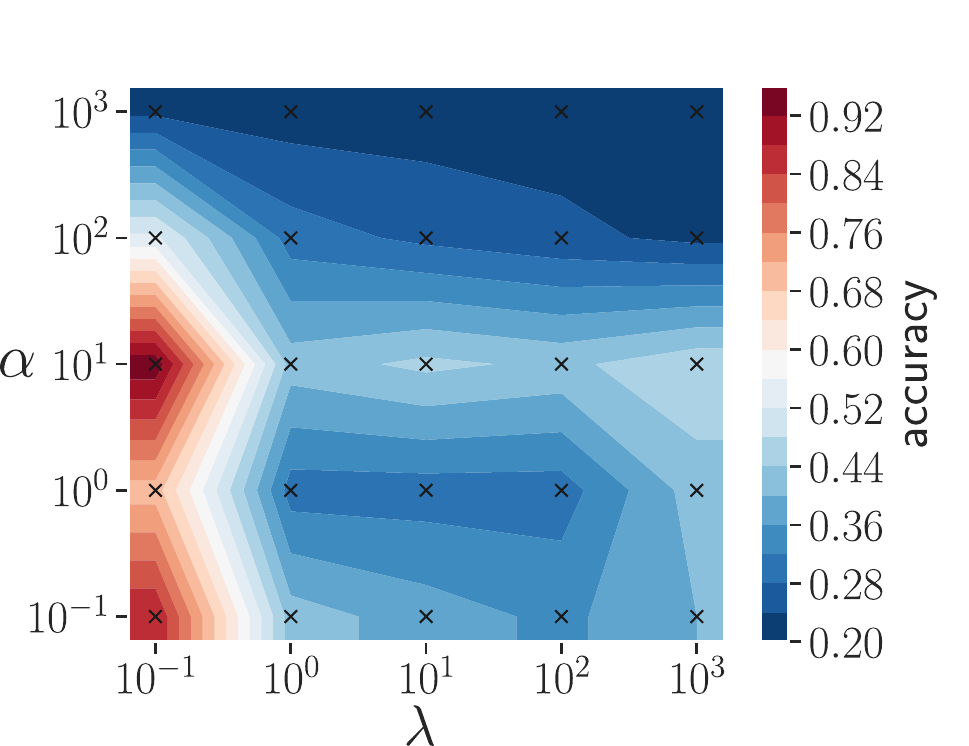}}&
    \subfloat{\includegraphics[trim=.0cm 0.cm .0cm 1.cm, clip,width=0.45 \columnwidth]{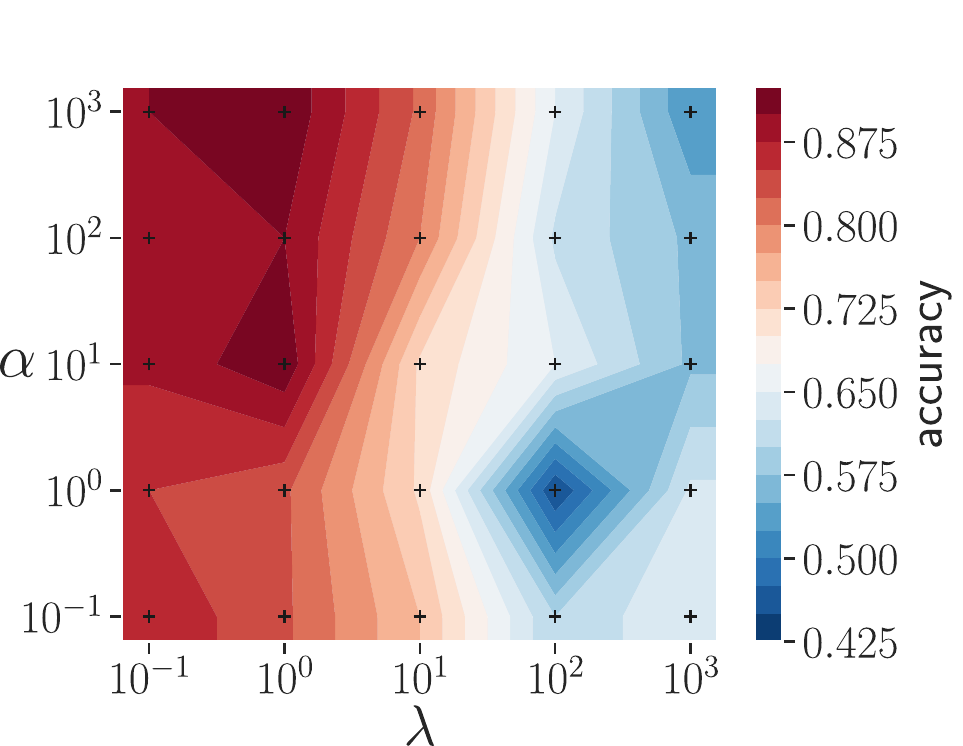}} \vspace{-2mm}\\
    \multirow{1}{*}[60pt]{\rotatebox[origin=c]{90}{\textbf{Cell-level}}} &
    \subfloat{\includegraphics[trim=.0cm 0.cm .0cm 1.cm, clip,width=0.45 \columnwidth]{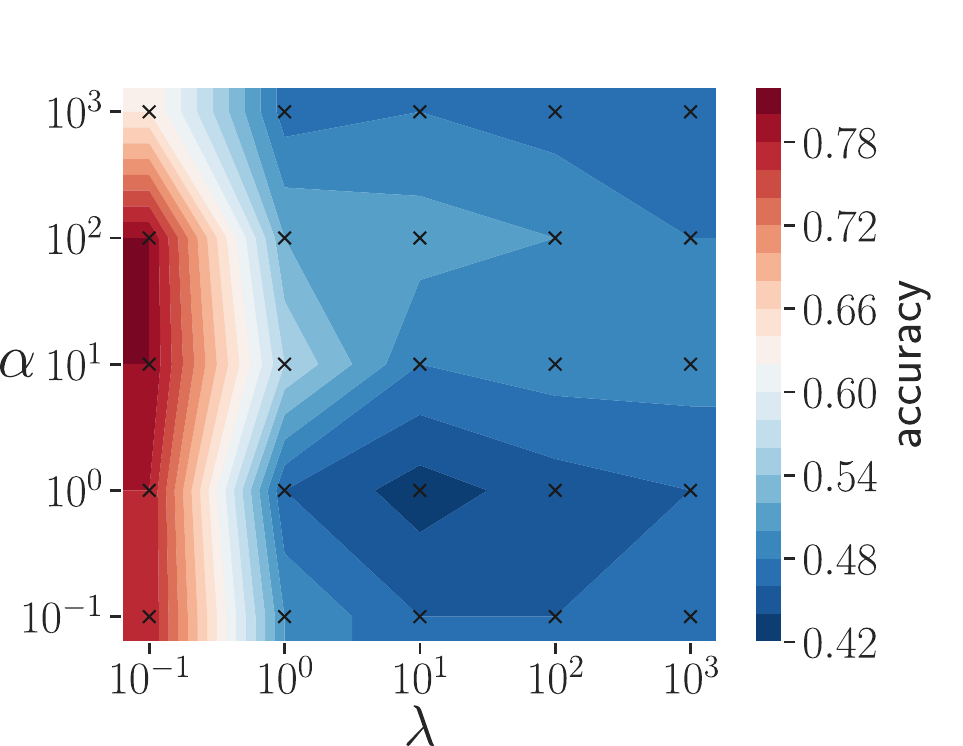}}&
    \subfloat{\includegraphics[trim=.0cm 0.cm .0cm 1.cm, clip,width=0.45 \columnwidth]{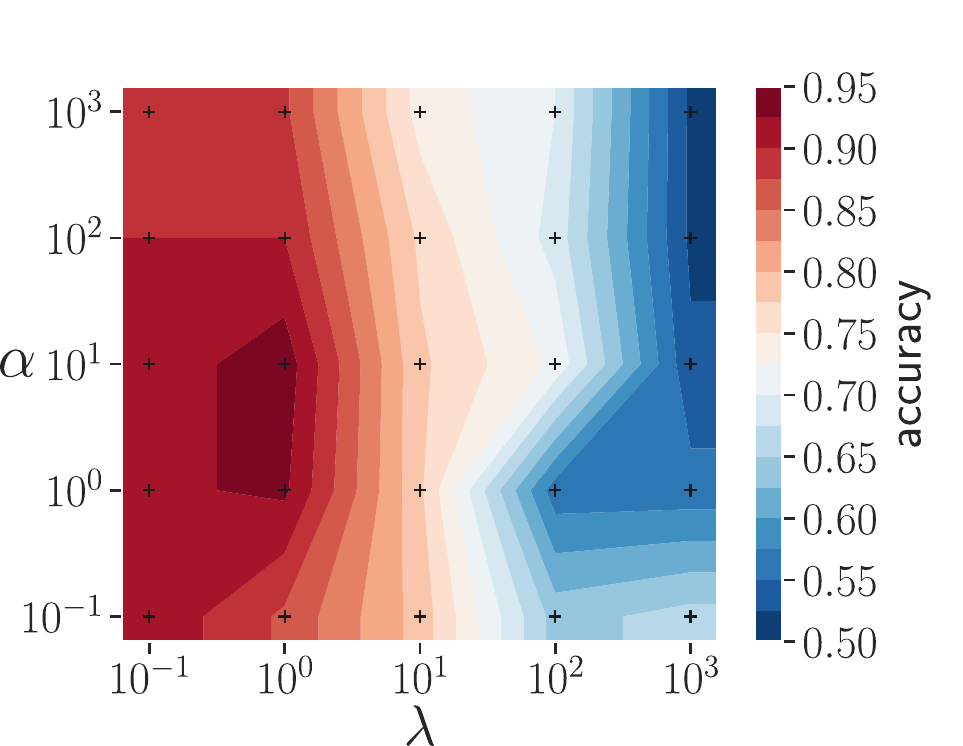}}\\
    \end{tabular}
    \caption{Grid search over hyperparameters $\alpha$ (margin) and $\lambda$ (regularization strength) on: (a) synthetic data and (b) myocardial infarction scRNA data. The classification accuracy on validation splits shows unique local maxima with smooth surfaces in the considered ranges which enables an efficient tuning.}
    \label{fig:hyper}
    \vspace{-0.3cm}
\end{figure}   

\subsection{Influence of Hyperparameters}
\label{sec:hyper}
As introduced in \autoref{sec:ggml}, the learning process of GGML is mainly influenced by hyperparameters $\alpha$ (margin) and $\lambda$ (regularization strength). Hyperparameters are tuned with a grid search on 20\% validation splits from the 10 test-train-validation splits in the classification task. \autoref{fig:hyper} shows the respective classification performance for the high-dimensional synthetic dataset and the scRNA dataset on myocardial infarction. For the synthetic data with few dimensions, a small L1 regularization strength ($\lambda < 1$) with moderate margins ($\alpha=10$) achieves optimal performance. For the real-world scRNA data with many correlated genes, optimization with an L1 norm becomes unstable. On the myocardial infarction dataset optimal results are achieved for a moderate L2 regularization ($\lambda=1$). As expected, too strong regularization leads to underfitting in all considered datasets.
In the noisy real-world datasets with large variances within classes (i.e. disease states) large margins $\alpha$ can lead to a better generalizability, and thus a higher classification accuracy in the validation split.

The hyperparameters $k$ and $t$ are strongly related to computational optimizations. $k$ specifies the maximal rank of the learned decomposed matrix $\widetilde{\bm{W}}$ of the Mahalanobis matrix. $t$ specifies the number of neighbors from each class that are considered to form triplets of relative relationships.  Setting these hyperparameters corresponds to a trade-off between computational complexity and quality of approximation. In the datasets presented, a 10-fold reduction in computation time is achieved while maintaining high classification performance. A more detailed ablation study w.r.t. computation times can be found in appendix \ref{tab:abla}. 

\section{RELATED WORK}
\label{sec:related_work}
This section differentiates our proposed framework from existing methods in Global and Ground Metric Learning. 
Global Metric Learning learns a single distance metric tuned to a particular task \cite{yang2006distance,kulis2013metric}. 
The metric is globally applicable to any two points as opposed to Local Metric Learning where multiple local metrics are learned with varying notions of locality \cite{dong2019learning}. 
In this paper, we consider approaches that learn a Mahalanobis distances between data points from (multiple) class labels.
Neighborhood Components Analysis (NCA)\cite{goldberger2004neighbourhood} and Large Margin Nearest Neighbor (LMNN) \cite{weinberger2009distance} maximize the margin between the data points from a class and differently labeled ones.
While these methods optimize over a decomposed matrix (see \autoref{eq:mahala}), Information Theoretic Metric Learning (ITML) \cite{davis2007information} is regularizing with the log-determinant divergence to enforce positive semi-definiteness. 
All of these global approaches fail to handle heterogeneous data from multimodal distributions. 
Existing work addresses this through learning multiple local metrics for each mode or other notions of locality \cite{wang2012parametric,dong2019learning}. Our method aims at learning a single metric that globally describes heterogeneous data. 

To achieve this, we consider a different methodological approach by learning global metrics as ground metrics, rather than increasing the model's complexity by means of Local or Deep Metric Learning. 
While both approaches are promising future directions in ground metric learning, we believe that current OT methods can be significantly improved by ground metric learning of simple metrics such as the Mahalanobis distance as demonstrated in this work.
A promising Deep Metric Learning to consider as part of future work is Neuronal Optimal Transport (NOT) which trains Input Convex Neural Networks to compute transport plans for an underlying ground metric \cite{korotin2022neural}. However, it is not a ground metric learning approach as it requires a fixed (weak) ground metric as a prior. Nonetheless, NOT has found promising applications on scRNA data \cite{bunne2023learning} which might further benefit from extending this approach with Ground Metric Learning.

The initially proposed Ground Metric Learning framework \cite{cuturi2014ground} learns a metric distance matrix between supports. It is not a learned metric function in a sense that it can compare two unseen data points. 
Rather, it learns specific distance values between supports of distributions that satisfy the metric property. A similar formulation was earlier introduced as supervised EMD in the context of computer vision \cite{wang2012supervised}.  
Another approach proposes the use of singular vectors to learn such distance values in an unsupervised manner \cite{huizing2022unsupervised}. 
While these approaches do not require prior knowledge about the relationships between supports, they also cannot leverage such information given by the supports position in some underlying space. 
Furthermore, by directly learning distances between supports, such supervised approaches cannot be applied to datasets without shared supports. For such datasets, the resulting metric distance matrix contains $n^2$ values for $n$ distinct supports which becomes under-determined and prone to overfitting. This is clearly the case for patient-level scRNA data where no identical cells are sampled for different patients which leads to disjoint supports. 
The Supervised Word Mover's Distance learns a Mahalanobis distance of word embeddings using an approach like NCA \cite{huang2016supervised}.
However, it also assumes shared supports in the form of histograms over a fixed vocabulary. 
Other recently proposed Ground ML methods learn geodesics to interpolate intermediate steps of trajectories \cite{scarvelis2022riemannian,kapusniak2024metric}. While these approaches yield promising results, they only apply to datasets with known timestamps which is generally not the case for disease progressions in patient-level scRNA data. 

\section{CONCLUSION}
We introduced a global ground metric learning framework, GGML, and demonstrated its strong performance and robustness using large synthetic and scRNA datasets from diverse diseases. 
The effectiveness of GGML in both classification and clustering of disease states underscores its capability to handle high-dimensional heterogeneous data. 
As the learned metrics generalize to unseen data, GGML has the potential to enhance performance in many existing optimal transport applications while minimizing the risk of overfitting.

Notably, GGML learns relationships at the level of distributions and data points.
For scRNA data, this means that GGML can accurately capture disease states of patients (distributions) and cells (points) by implicitly learning relevant directions (subspaces) within the gene space along which distances best discriminate between classes. 
Using a low rank approximation, our method efficiently learns a low dimensional subspace with axes corresponding to distinct gene activation patterns.
Weights in the learned subspace can be directly interpreted as gene importance, identifying disease related genes in all considered diseases.
To illustrate this, we have provided an exemplary analysis of the feature importance and found results matching existing literature on the respective diseases.  
However, our framework is not limited to scRNA data. GGML has the potential to generate new insights in various other domains where optimal transport is used with heterogeneous data.

\newpage
\subsection*{Acknowledgments}
This work was funded as part of the Graphs4Patients Consortia by the BMBF (Federal Ministry of Education and Research) and the Ministry of Culture and Science of North Rhine-Westphalia (NRW R\"uckkehrprogramm). The authors thank K\"ubra G\"uven and Simon Jonas for their technical work on distributing the code as a Python package.

\bibliography{ggml.bib}

\newpage
\section*{Checklist}
 \begin{enumerate}

 \item For all models and algorithms presented, check if you include:
 \begin{enumerate}
   \item A clear description of the mathematical setting, assumptions, algorithm, and/or model. \textbf{Yes}, see \autoref{sec:background}, \autoref{sec:ggml} and \autoref{sec:comp}.
   \item An analysis of the properties and complexity (time, space, sample size) of any algorithm. \textbf{Yes}, reducing the cubic scaling of optimized terms to linear \autoref{sec:ggml}, and a low rank approximation of the learned Mahalanobis distance in \autoref{sub:low_rank_mahala}. Results on computational improvements can be found in \autoref{tab:abla}.
   \item (Optional) Anonymized source code, with specification of all dependencies, including external libraries. \textbf{Yes} 
 \end{enumerate}

 \item For any theoretical claim, check if you include:
 \begin{enumerate}
   \item Statements of the full set of assumptions of all theoretical results. \textbf{Yes}
   \item Complete proofs of all theoretical results. \textbf{Yes}, in \autoref{sec:proof}.
   \item Clear explanations of any assumptions. \textbf{Yes}, in \autoref{sec:ggml}.    
 \end{enumerate}

 \item For all figures and tables that present empirical results, check if you include:
 \begin{enumerate}
   \item The code, data, and instructions needed to reproduce the main experimental results (either in the supplemental material or as a URL). \textbf{Yes}, see \href{https://www.github.com/DaminK/GlobalGround-MetricLearning}{github.com/DaminK/GlobalGround-MetricLearning}.
   \item All the training details (e.g., data splits, hyperparameters, how they were chosen). \textbf{Yes}, in \autoref{sec:applications} and \autoref{sec:comp}
         \item A clear definition of the specific measure or statistics and error bars (e.g., with respect to the random seed after running experiments multiple times). \textbf{Yes}, in \autoref{sec:applications} and \autoref{sec:results2}
         \item A description of the computing infrastructure used. (e.g., type of GPUs, internal cluster, or cloud provider). \textbf{Yes}, in \autoref{sec:comp}
 \end{enumerate}

 \item If you are using existing assets (e.g., code, data, models) or curating/releasing new assets, check if you include:
 \begin{enumerate}
   \item Citations of the creator If your work uses existing assets. \textbf{Yes}, used packages, notably Python Optimal Transport (POT) and PyTorch, are cited.
   \item The license information of the assets, if applicable. \textbf{Yes}, MIT for POT and BSD-3 for PyTorch.
   \item New assets either in the supplemental material or as a URL, if applicable. \textbf{Not Applicable}.
   \item Information about consent from data providers/curators. \textbf{Yes}, we used published and publicly available scRNA datasets from CEllxGENE.
   \item Discussion of sensible content if applicable, e.g., personally identifiable information or offensive content. \textbf{Not Applicable}.
 \end{enumerate}

 \item If you used crowdsourcing or conducted research with human subjects, check if you include:
 \begin{enumerate}
   \item The full text of instructions given to participants and screenshots. \textbf{Not Applicable}.
   \item Descriptions of potential participant risks, with links to Institutional Review Board (IRB) approvals if applicable. \textbf{Not Applicable}.
   \item The estimated hourly wage paid to participants and the total amount spent on participant compensation. \textbf{Not Applicable}.
 \end{enumerate}

 \end{enumerate}

\newpage
\onecolumn

\hsize\textwidth
\linewidth\hsize \toptitlebar {\centering
{\Large\bfseries SUPPLEMENTARY MATERIALS \par}}
\bottomtitlebar \vskip 0.2in plus 1fil minus 0.1in

\setcounter{section}{0}
\renewcommand{\thesection}{\Alph{section}}

\begin{figure}[b!]
  \resizebox{0.98\linewidth}{!}{
  \centering
  \setlength{\tabcolsep}{1pt}
  \renewcommand{\arraystretch}{0.1}
  \begin{tabular}{lcccccccc}
  & (a) Euclidean & (b) Manhatten & (c) Cosine & (d) LFDA & (e) ITML & (f) \textbf{GGML}  \vspace{-2mm}\\
  \multirow{1}{*}[60pt]{\rotatebox[origin=t]{90}{\textbf{Patient-level}}} &
  \subfloat{\includegraphics[trim=.0cm 0.0cm .cm 0.cm, clip, width=0.2 \textwidth]{./figures/scRNA/breastcancer/umap_euc_patients_disease.png}} &
  \subfloat{\includegraphics[trim=.0cm 0.0cm .cm 0.cm, clip, width=0.2 \textwidth]{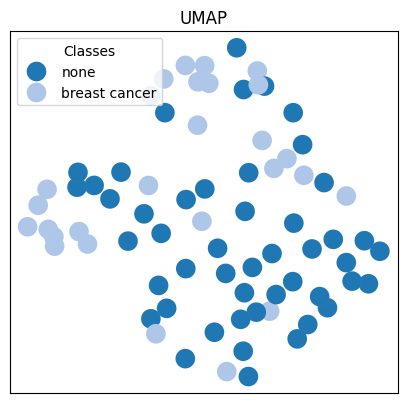}}& 
  \subfloat{\includegraphics[trim=.0cm 0.0cm .cm 0.cm, clip, width=0.2 \textwidth]{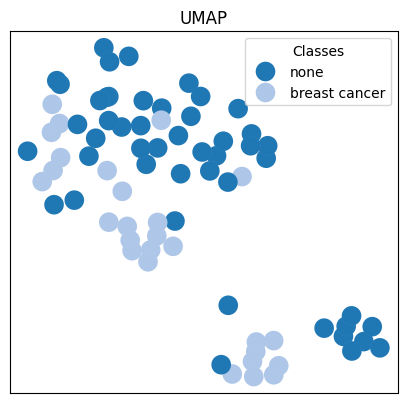}}& 
  \subfloat{\includegraphics[trim=.0cm 0.0cm .cm 0.cm, clip, width=0.2 \textwidth]{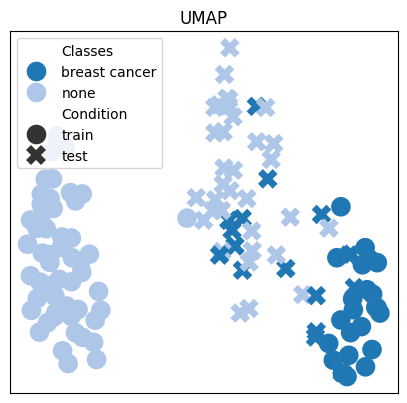}}&  
  \subfloat{\includegraphics[trim=.0cm 0.0cm .cm 0.cm, clip, width=0.2 \textwidth]{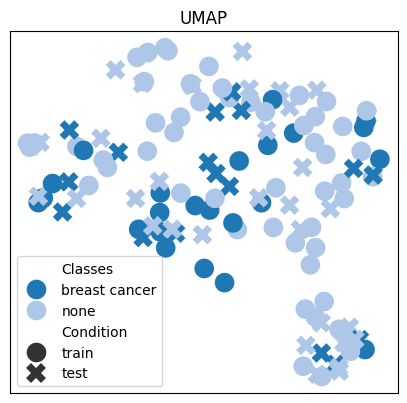}}&
  \subfloat{\includegraphics[trim=.0cm 0.0cm .cm 0.cm, clip, width=0.2 \textwidth]{./figures/scRNA/breastcancer/umap_ggml_patient.png}}\\
  \multirow{1}{*}[100pt]{\rotatebox[origin=B]{90}{\textbf{Cell-level}}} &%
  \subfloat{\includegraphics[trim=.0cm 0.0cm .cm 0.cm, clip, width=0.2 \textwidth]{./figures/barycenter/euc_disease.png}}&
  \subfloat{\includegraphics[trim=.0cm 0.0cm .cm 0.cm, clip, width=0.2 \textwidth]{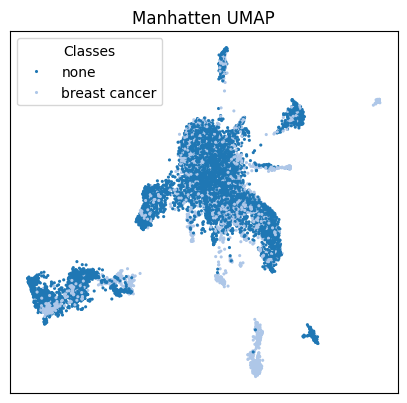}}&
  \subfloat{\includegraphics[trim=.0cm 0.0cm .cm 0.cm, clip, width=0.2 \textwidth]{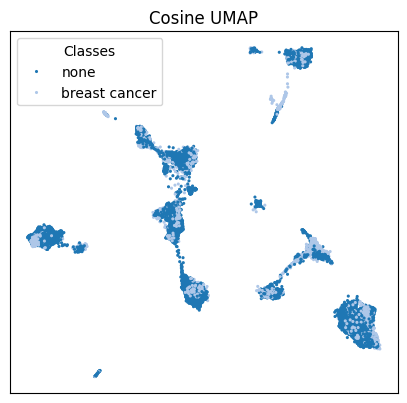}}&
  \subfloat{\includegraphics[trim=.0cm 0.0cm .cm 0.cm, clip, width=0.2 \textwidth]{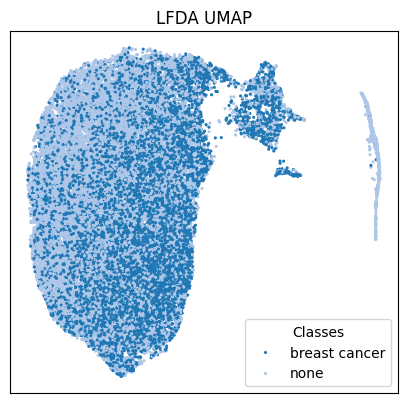}}&
  \subfloat{\includegraphics[trim=.0cm 0.0cm .cm 0.cm, clip, width=0.2 \textwidth]{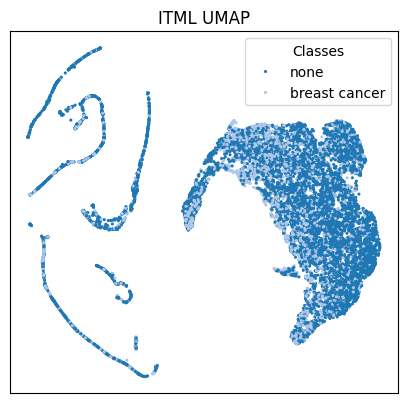}}&
  \subfloat{\includegraphics[trim=.0cm 0.0cm .cm 0.cm, clip, width=0.2 \textwidth]{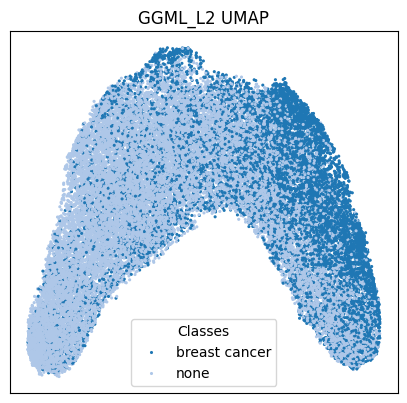}} \\
  \end{tabular}
  }
  \caption{Embeddings of patients and cells for scRNA-seq data from different diseases using $d_\theta$ learned by Global Ground Metric Learning (GGML) and Euclidean $d_2$ as baseline. To highlight the capabilities of GGML to generalize to unseen data the shown plots are produced by only learning on half of the data points as indicated.
  Relative weights of $\theta$ can be directly interpreted as gene importance in distinguishing disease stages. }
  \label{fig:appendix_embs}
\end{figure}
\section{RESULTS}
\label{sec:results2}
\subsection{Embeddings}
\label{ssec:emb}
\autoref{fig:appendix_embs} shows the patient-level and cell-level embeddings of all competing methods on the breast cancer scRNA dataset. Methods are trained as before on a train split of the data. For visualization purposes the learned (and unsupervised) metrics are shown on all patients. We also show the embedding of all  did not terminate on this ds as a baseline. \autoref{fig:scRNA_results} only showed cells that were correctly classified with at least 90\% across test-splits. This indicates whether cells are distinguishable between disease stages and served as proxy for being involved in the different disease stages.
NCA and LMNN did not terminate on this dataset due to time or memory constraints as seen in \autoref{tab:clust}. 

\subsection{Clustering}
\label{ssec:clust}
This section contains detailed clustering results on the presented datasets shown in \autoref{tab:clust}. Patient-level clusters corresponded to the different disease states of the tissue samples. On the cell-level, cells may form disease-related sub-clusters while unrelated cells might mix into additional clusters. To account for this, we do not enforce a fixed number of clusters for agglomerative hierarchical clustering at the cell level. Instead, we aggregate clusters until reaching a threshold based on the median distance of all pairwise global distances in the dataset.
For synthetic data, the number of clusters is predetermined by construction at both levels. To establish a baseline with reliable ground truth, the two non-differentiable modes (corners in Figure 2(a)) are excluded and only the modes that can be differentiated are clustered. The results shown here use parameters $\alpha=1,\lambda=1$ for the synthetic data and $\alpha=100,\lambda=100$ for the scRNA data and do not reflect the hyperparameter tuning for the classification.

The clustering is evaluated using the Mutual Information (MI), Adjusted Rand Index (ARI), and Variational Information (VI). The MI quantifies the shared information between true labels $U$ and cluster $V$ assignments by computing $MI(U;V)=\sum_{i,j} P(i,j) \log \frac{P(i,j)}{P(i)P(j)} $, where $P(i,j)$ is the joint probability distribution of the clusters. ARI measures the ratio of pairs of samples that are correctly classified, adjusted for random chance. It is given as $ARI= \frac{RI -E[RI]}{max(RI) -E [RI]} $, where $E[RI]$ is the expected value of the Rand Index (RI). The optimal ARI value is 1 for perfect clusterings w.r.t labels. VI can be calculated as $VI(U,V)=H(U)+H(V) - 2MI(U;V)$, where $H$ is the entropy of the resp. clustering. VI is a distance measure between clusterings where 0 corresponds to perfect clusterings.

The results in \autoref{tab:clust} demonstrate that GGML outperforms other methods across all datasets at the patient level and achieves better overall performance at the cell level. Cell-level clustering is more challenging for all methods, as expected, due to the inclusion of unrelated cell types and the formation of subcluster among disease-related cell types.  These subcluster are not well captured by the clustering metrics when evaluated against patient disease states, which we used as a proxy due to the lack of ground truth distances for cell-level disease states.

\begin{table}
  \centering
  \resizebox{0.85\linewidth}{!}{
  \footnotesize
  \setlength{\tabcolsep}{2pt}
  \begin{tabular}{lr||ccc|ccc|ccc|ccc|ccc}
  
  &  &  \multicolumn{3}{c|}{$\text{Synth}_{2D}$} & \multicolumn{3}{c|}{$\text{Synth}_{200D}$} & \multicolumn{3}{c|}{Kidney disease} & \multicolumn{3}{c|}{Breast cancer} & \multicolumn{3}{c}{Myocard. inf.}\\
  \multirow{1}{*}{Method}  &  & MI & ARI & VI & MI & ARI & VI& MI & ARI & VI & MI & ARI & VI & MI & ARI & VI\\
  \midrule
  Euclidean & \multirow{8}{*}{\rotatebox[origin=c]{90}{\it{patient-level}}} & 0.00 & -0.07 & 2.19 & 0.00 & -0.07 & 2.13 & 0.07 & 0.02 & 1.11 & 0.01 & -0.02 & 0.68 & 0.11 & 0.02 & 1.21 \\
  Manhatten & & 0.00 & -0.07 & 2.13 & 0.00 & -0.07 & 2.19 & 0.07 & 0.02 & 1.11 & 0.03 & 0.13 & 0.93 & 0.22 & 0.17 & 1.00 \\
  Cosine & & 0.03 & -0.03 & 1.97 & 0.07 & -0.00 & 1.24 & 0.07 & 0.02 & 1.11 & 0.01 & -0.02 & 0.68 & 0.07 & 0.00 & 1.19 \\
  LMNN & &  0.01 & -0.07 & 2.17 & 0.01 & -0.06 & 2.17 & \multicolumn{3}{c|}{\textit{OOM}} & \multicolumn{3}{c|}{\textit{OOM}}& \multicolumn{3}{c}{\textit{OOM}} \\
  LFDA & & 0.03 & -0.05 & 2.07 & 0.08 & -0.01 & 1.40 & 0.05 & -0.01 & 1.39 & 0.06 & 0.13 & 0.68 & 0.64 & 0.53 & 0.75 \\
  NCA & &  0.01 & -0.06 & 2.15 & 0.04 & -0.03 & 2.10 & \multicolumn{3}{c|}{\textit{OOT}} & \multicolumn{3}{c|}{\textit{OOT}}& 0.87 & 0.77 & 0.37  \\
  ITML & & 0.18 & 0.02 & 1.68 & 0.22 & 0.11 & 1.38 & 0.06 & 0.01 & 1.27 & 0.01 & -0.02 & 0.68 & 0.09 & 0.03 & 1.15 \\
  GGML & & \textbf{1.10} & \textbf{1.00} &\textbf{ 0.00} &\textbf{ 1.10 }& \textbf{1.00} &\textbf{ 0.00 }&  \textbf{0.65} & \textbf{0.66} & \textbf{0.60} & \textbf{0.21} & \textbf{0.49 }& 0.79 & \textbf{0.91} & \textbf{0.92} & \textbf{0.20} \\
  \midrule
  Euclidean & \multirow{8}{*}{\rotatebox[origin=c]{90}{\it{cell-level}}} &0.00 & -0.00 & 1.78 & 0.00 & -0.00 & 1.79 & 0.18 & 0.00 & 3.17 & \textbf{0.11} & -0.00 & 2.87 & 0.29 & 0.11 & 2.83 \\
  Manhatten & & 0.00 & -0.00 & 2.18 & 0.00 & -0.00 & 1.79 & 0.15 & 0.01 & 3.12 & 0.06 & 0.02 & 2.33 & 0.27 & 0.11 & 2.25 \\
  Cosine & & 0.05 & 0.02 & 2.03 & 0.34 & 0.01 & 3.05 & 0.18 & 0.00 & 3.09 &\textbf{ 0.11} & 0.03 & 2.91 & 0.33 & 0.09 & 3.40 \\
  LMNN & & 0.01 & 0.01 & 2.15 & 0.00 & 0.00 & 1.77 & \multicolumn{3}{c|}{\textit{OOM}} & \multicolumn{3}{c|}{\textit{OOM}}& \multicolumn{3}{c}{\textit{OOM}}  \\
  LFDA & & 0.06 & 0.03 & 1.87 & 0.00 & -0.00 & 2.40 & 0.04 & -0.01 & 2.74 & 0.04 &\textbf{ 0.06} & 2.47 & 0.51 & 0.64 & 0.71 \\
  NCA & & 0.00 & 0.00 & 2.16 & 0.10 & 0.06 & 1.93  & \multicolumn{3}{c|}{\textit{OOT}} & \multicolumn{3}{c|}{\textit{OOT}}& 0.23 & 0.04 & 1.50 \\
  ITML & & 0.15 & 0.09 & 1.82 & 0.01 & 0.00 & 2.46 & 0.04 & 0.00 & 2.47 & 0.02 & 0.04 & 2.27 & 0.07 & 0.01 & 2.77 \\
  GGML & & \textbf{0.77} & \textbf{0.75} & \textbf{0.65} & \textbf{0.68} & \textbf{0.64} & \textbf{ 0.83} &\textbf{ 0.24 }& \textbf{0.22} &\textbf{ 1.67} & 0.06 & \textbf{0.06} & \textbf{1.61} & \textbf{0.63} & \textbf{0.71} & \textbf{0.81} \\
  \end{tabular}}
  \caption{Clustering performance of all considered metrics on the patient- and cell-level, evaluated against patient disease-states as proxy for an unknown ground truth. GGML consistently outperforms other methods, as measured by Mutual Information (MI), Adjusted Rand Index (ARI), and Variation of Information (VI).}
  \label{tab:clust}
\end{table}

\begin{figure}
  \centering
      \resizebox{1.0\linewidth}{!}{
      \begin{tabular}{ccc}
      \multicolumn{3}{c}{
      \begin{tabular}{cc}
      \subfloat[Axis 1]{\includegraphics[height=7cm]{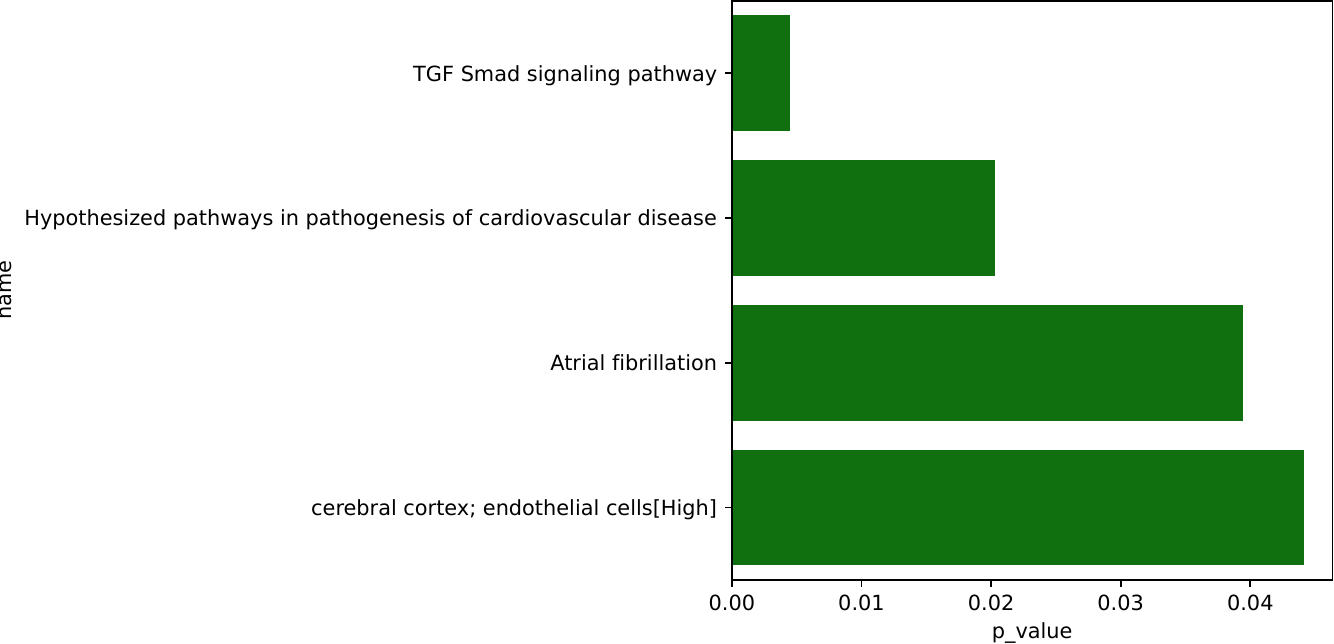}}&
      \subfloat[Axis 2]{\includegraphics[height=7cm]{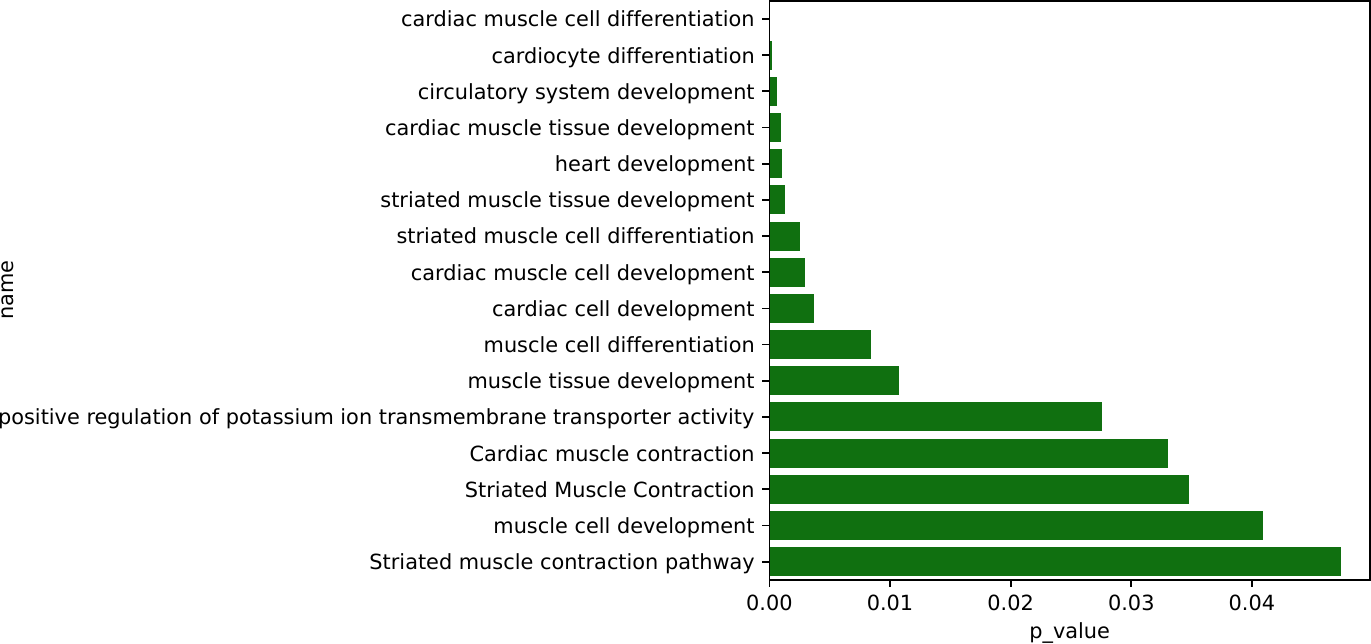}}
      \end{tabular}} \\ \vspace*{-0.3cm} 
      \subfloat[Axis 3]{\includegraphics[height=7cm]{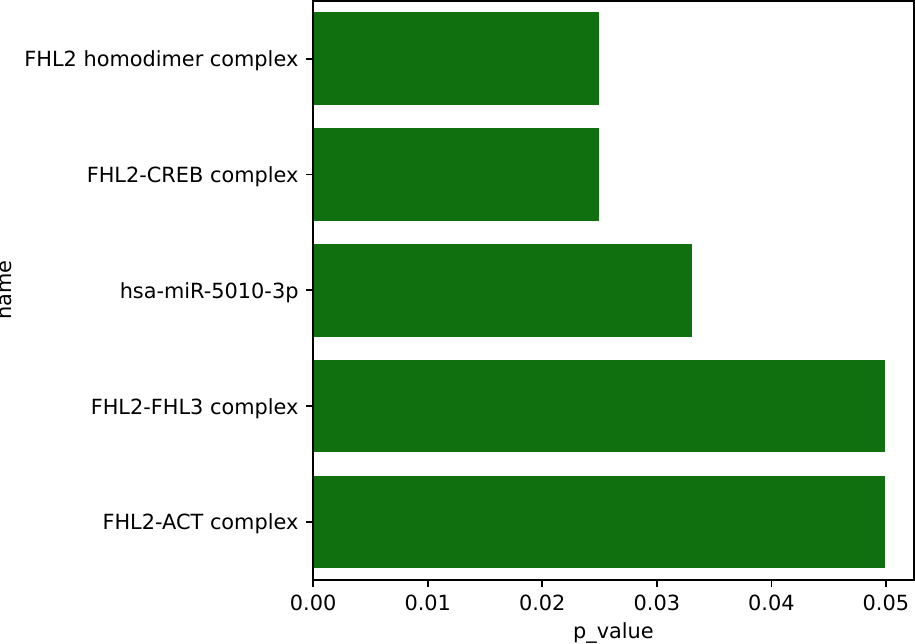}}&
      \subfloat[Axis 4]{\includegraphics[height=7cm]{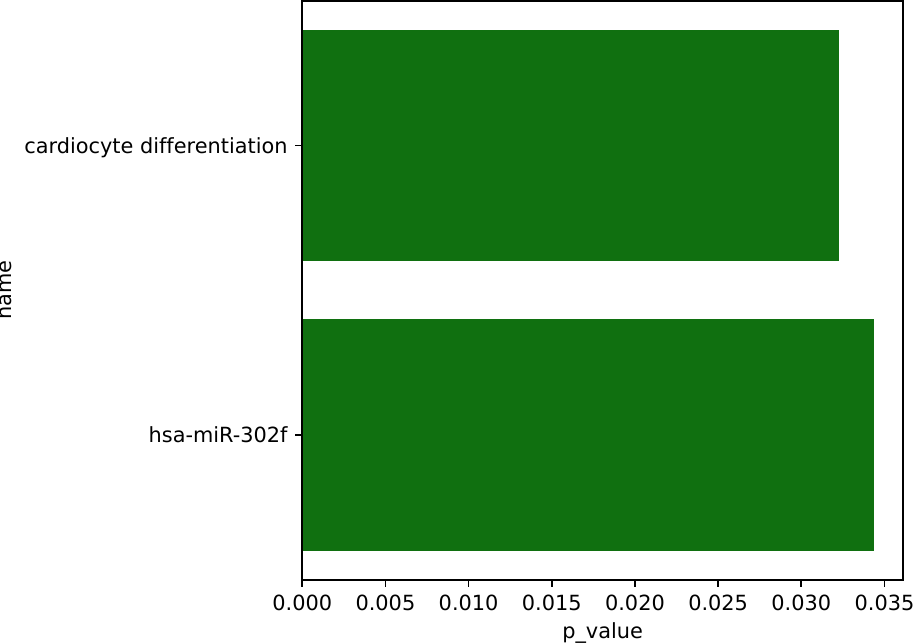}}&    \subfloat[Axis 5]{\includegraphics[height=7cm]{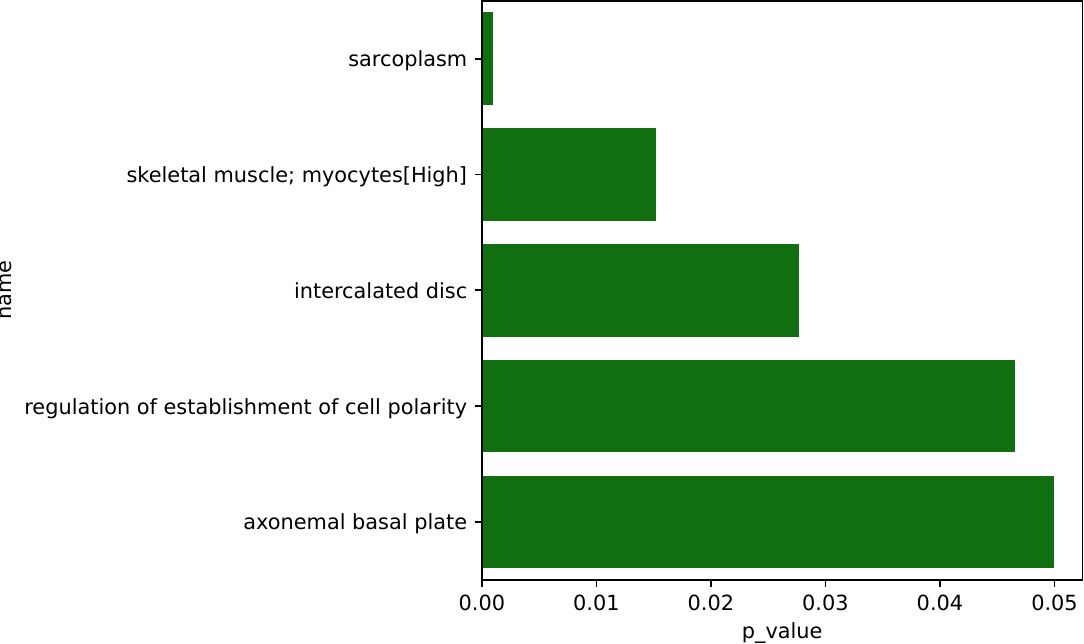}}%
      \end{tabular}}
      \caption{Significantly enriched biological processes for the individual axes of the learned subspace in the myocardial infarction dataset. The processes highlight distinct and relevant pathways, demonstrating GGML's ability to uncover disease-related gene expression changes using only patient disease states.}    
      \label{fig:biological_process}
      \vspace*{-0.2cm}
\end{figure}

\subsection{Biological Processes in learned Subspaces}
\label{ssec:gene_enrich}
The individual rows of $\widetilde{W}$ correspond to different axis of the learned subspace. The axes can be directly interpreted as distinct genes linked to different disease-related processes. We present an exemplary analysis of the learned subspace for the myocardial infarction dataset, highlighting numerous relevant processes. Notably, all results were derived using only single-cell gene expression data and patient-level disease labels for training GGML.

To identify biological processes from the most important genes within each axis, we performed gene enrichment analysis. Identified genes were mapped against existing databases of functionally related genes and their corresponding processes using g:Profiler \cite{raudvere2019g}. \autoref{fig:biological_process} shows the significantly enriched processes for each axis.

Each axis contains distinct and interpretable processes relevant to myocardial infarction. For instance, axis 2 captures various heart muscle cell differentiation and development processes that drive disease progression, while axis 1 includes a process that is already hypothesized to drive cardiovascular diseases. These findings highlight GGML's promising capabilities and interpretability in identifying disease-related gene expression changes and associated biological processes.

\section{Proofs}
\label{sec:proof}
\setcounter{thm}{0}
\begin{thm} 
The GGML loss is 0 if, and only if, the global ground metric $d_\theta$ in $W_\theta$ approximates the ground truth distances $W^*$ with margin $\alpha$. 
$$\mathcal{L}_\alpha(\theta,X,\tilde{\mathcal{T}_t})\!=\!0 \: \text{iff} \: \:  W_\theta \approx_\alpha W^*$$ 
\end{thm} 
We show that both sides are equivalent for $\alpha \in \mathbb{R}_{\geq0}$. 
\begin{align*}
    & W_\theta \approx_\alpha W^* \\
    \Leftrightarrow \quad & \forall(i,j,k)\in\tilde{\mathcal{T}}_t : W_\theta(X_j,X_k) - W_\theta(X_i,X_j) \geq  \alpha \\
    \Leftrightarrow \quad & \forall(i,j,k)\in\tilde{\mathcal{T}}_t : \\
    & W_\theta(X_i,X_j) - W_\theta(X_j,X_k) + \alpha \leq 0 \\
    \Leftrightarrow \quad & \forall(i,j,k)\in\tilde{\mathcal{T}}_t : \\
    & \max \left( W_\theta(X_i,X_j) - W_\theta(X_j,X_k) + \alpha, 0 \right) = 0 \\    
    \Leftrightarrow \quad & \forall(i,j,k)\in\tilde{\mathcal{T}}_t : \mathcal{L}_\alpha(\theta,X,(i,j,k)) = 0 \\
    \Leftrightarrow \quad & \sum\limits_{t \in \tilde{\mathcal{T}}_t} \mathcal{L}_\alpha(\theta,X,t) =  \mathcal{L}_\alpha(\theta,X,\tilde{\mathcal{T}}_t) =0 \\
\end{align*}
The last equivalence holds true as $\mathcal{L}_\alpha(\theta,X,t) \geq 0$.\\

\begin{thm}[Triplet Loss Bound]
For all distributions $X$ and triplets $(i,j,k)\in\tilde{\mathcal{T}}_t$, there exist a $\theta$ such that $\mathcal{L}(\theta,X,(i,j,k))$ 
is at most $\alpha$. 

For all distributions $X$ and sets of triplets $\tilde{\mathcal{T}}_t$, the minimal loss $\mathcal{L}(\theta,X,\tilde{\mathcal{T}_t})$ is at most $\alpha  |\tilde{\mathcal{T}_t}|$. 
\end{thm}
Let $d_\mymathbb{0}: \Omega^2 \mapsto 0$ be the zero function that maps pairs of points $x,y\in\Omega$ to a distance of $0$. With $d_\mymathbb{0}$ as ground metric, it holds that $W_\mymathbb{0}(X_i,X_j)=0$ for any distributions $X_i,X_j,$ as $d_\mymathbb{0}(x,y)=0$ for all points $x,y$. Clearly, it also holds for all triplets $\forall (i,j,k)\in\tilde{\mathcal{T}}_t$ that $W_\mymathbb{0}(X_i,X_j) - W_\mymathbb{0}(X_j,X_k)=0$.

Consider that the zero matrix $\mymathbb{0}$ is PSD and thus a valid covariance matrix of the Mahalanobis distance. Hence, the zero function $d_\mymathbb{0}$ is in the hypothesis space of learnable Mahalanobis ground metrics. It follows that:
\begin{align}
&(i,j,k)\in\tilde{\mathcal{T}}_t:\min_\theta W_\theta(X_i,X_j) - W_\theta(X_j,X_k) \leq 0  \nonumber \\ 
\implies \quad &(i,j,k)\in\tilde{\mathcal{T}}_t:\min_\theta\mathcal{L}_\alpha(\theta,X,(i,j,k)) \leq \alpha    \nonumber
\end{align}

with the equality being satisfied for $\theta=\mymathbb{0}$. 

Given that $\exists\theta:\mathcal{L}(\theta,X,t) \leq \alpha$ and using the same reasoning as above, it follows:
$$\min\limits_\theta\sum_{t\in\tilde{\mathcal{T}_t} } \mathcal{L}(\theta,X,t) \leq  \min\limits_\theta\mathcal{L}(\theta,X,t) \, |\tilde{\mathcal{T}_t}|  \leq \alpha \, |\tilde{\mathcal{T}_t}|$$
with equality given for $\theta=\mymathbb{0}$.

\section{Implementation \& Computation}
\label{sec:comp}

\subsection{Neighborhood Parameters}
In the context of using a kNN classifier with metric learning approaches, multiple neighborhood parameters arise for which we want to provide some clarification.
For global metric learning, it refers to the neighbors of a data point, respective distribution for ground metric learning. 
We have introduced a neighborhood parameter $t$ for our global ground metric learning on the distribution level determining on how many neighbors are used to train on. Refer to \autoref{sec:ggml} for details on how this parameter is used to construct the triplet sets containing relative relationships between distributions. Due to training on only half of the data and the presence of small classes in scRNA disease states, we have set $t=3$. For the synthetic data, where classes are larger, we use $t=5$.
In the context of kNN classification, the k-closest neighbors predict the label of a data point or distribution. We used $k=5$ for patient-level classification and $k=100$ cell-level classification due to the significant differences in number of data points and expected heterogeneity. Note that in all other contexts we refer to the rank of the decomposed Mahalanobis matrix with $k$.

\subsection{Packages}
We use Python Optimal Transport ~\cite{flamary2021pot} and PyTorch \cite{paszke2019pytorch} to compute a differentiable Wasserstein distance w.r.t. to the parameter of the underlying ground metric. ADAM \cite{kingma2014adam} is used to optimize the loss function. Competing metric learning methods are taken from metric-learn \cite{de2020metric}. Computation of fixed metrics and the evaluation of the classification and clustering benchmark is done with Scikit-learn \cite{pedregosa2011scikit}. Gene enrichment is performed with g:Profiler \cite{raudvere2019g}. 

\subsection{Computing Infrastructure \& Training Details}
All results were computed on an internal computing node running Linux, equipped with 64 cores (AMD EPYC 7763 at 3.5 GHz) and 1 TB RAM. We train GGML using the ADAM optimizer with a learning rate of 0.01. The training is conducted on minibatches that contain 128 triplets. Training is stopped after 30 to 50 iterations depending on the size of the dataset. Each epoch takes approximately 130 seconds ($\text{Synth}_{200D}$) to 950 seconds (breast cancer) to compute for rank $k=5$ and $t=3$ neighbors, as indicated in \autoref{tab:abla}. The breast cancer dataset is the largest considered dataset with 131 patients and 714k cells which are subsampled to 131k cells in our experiments.

In the benchmarking pipeline, we run competing methods subsequently so that methods have access to the same resources. We suspended the training of a metric learning method after 8 hours of computation ("Out of time") or when trying to allocate more than 1TB of RAM ("Out of memory"). To improve computation time and make the benchmark more comparable, we perform a low-rank approximation with a pivoted Cholesky decomposition \cite{harbrecht2012low} for competing methods who can only learn full-rank Mahalanobis matrices $M$. 
As not all competing methods utilize multi-threading in their provided implementations, we refrained from discussing computation times as an indicator of the method's computational efficiency.

\begin{table}
\footnotesize
\centering
\setlength{\tabcolsep}{1pt}
\begin{tabular}{l|ccccc} 
\multicolumn{6}{c}{Panel A: $\text{Synth}_{200D}$} \\
\backslashbox{Neigh.}{Rank.} & \textbf{5}	&10	&50	&100	&200(full)\\
\toprule
1	          &12.9	&14.5	&15.8	&17.0	&22.3 \\
3	  &130.4	&132.5	&143.3	&156.0	&195.8\\
\textbf{5}	          &348.6	&361.3	&365.2	&417.3	&525.4\\
7	          &646.0	&687.6	&789.3	&866.7	&1040.5\\
9(all)	&1049.1	&1119.8	&1249.2	&1446.2	&1703.2\\
\end{tabular}
\begin{tabular}{l|cccccc}
\multicolumn{7}{c}{Panel B: breast cancer} \\
  \backslashbox{Neigh.}{Rank.} & \textbf{5} & 10 & 100 & 500 & 1000 & 8433(full)\\
  \midrule
  1 & 148.9 & 145.1 & 189.0 & 276.2 &  572.0 & OOM \\
  2 & 390.4 & 473.8 & 666.2 & 1256.9 & 3009.9 & OOM\\
  \textbf{3} & 955.4 & 1035.8 & 1548.8 & 2610.7 & 5877.5 & OOM\\
  4 & 1776.1 & 1924.6 & 2563.5 & 4665.4 & 10425.0 & OOM\\
  5 & 2587.5 & 3097.7 & 3519.5 & 7353.1 & 19159.3 & OOM \\
  
\end{tabular}
\caption{Ablation Study on computational improvements using a low rank approximation and a fixed numbers of neighbors, measured by average training time per epoch (in seconds). Parameters used in the results section of the paper are indicated in bold.}
\label{tab:abla}
\end{table}

\newpage

\end{document}